\documentclass[sigconf, screen]{acmart}
\makeatletter
\renewcommand\@formatdoi[1]{\ignorespaces}
\makeatother

\usepackage{booktabs} 
\usepackage{hyperref}
\usepackage{url}
\usepackage{graphicx}  
\graphicspath{{images/}}
\usepackage{subcaption, amsfonts, dcolumn}
\usepackage{amsmath}
\usepackage{float}
\usepackage{amssymb}
\usepackage{textcomp}
\usepackage[export]{adjustbox}
\usepackage{paralist}
\newcolumntype{d}[1]{D..{#1}}
\definecolor{gainsboro}{rgb}{0.9, 0.9, 0.9}
\usepackage{subcaption, caption}
 \usepackage{pstricks}
 \usepackage{epsfig}
 \usepackage[space]{grffile} 
 \usepackage{etoolbox} 
 
 \makeatletter 
 \patchcmd\Gread@eps{\@inputcheck#1 }{\@inputcheck"#1"\relax}{}{}
 \makeatother

\setcopyright{usgovmixed}

\begin{document}
\title[Network Signatures from Image Representation of Adjacency Matrices]{Network Signatures from Image Representation of Adjacency Matrices: Deep/Transfer Learning for Subgraph Classification}

\author{Kshiteesh Hegde}
\orcid{0000-0003-0839-940X}
\affiliation{%
  \institution{Rensselaer Polytechnic Institute}
  \streetaddress{110 8th St}
  \city{Troy}
  \state{NY}
  \postcode{12180}
}
\email{hegdek2@rpi.edu}

\author{Malik Magdon-Ismail}
\affiliation{%
  \institution{Rensselaer Polytechnic Institute}
  \streetaddress{110 8th St}
  \city{Troy}
  \state{NY}
  \postcode{12180}
}
\email{magdon@rpi.edu}

\author{Ram Ramanathan}
\authornote{Work was done when the author was at Raytheon BBN}
\affiliation{%
  \institution{Raytheon BBN Technologies}
  \streetaddress{Network Research Department}
  \city{Cambridge} 
  \state{MA} 
  \postcode{02138}
}
\email{ramanathan@bbn.com}

\author{Bishal Thapa}
\affiliation{%
  \institution{Raytheon BBN Technologies}
  \streetaddress{Network Research Department}
  \city{Cambridge} 
  \state{MA} 
  \postcode{02138}
}
\email{thapa@bbn.com}

\renewcommand{\shortauthors}{K. Hegde et al.}

\begin{abstract}
  We propose a novel subgraph image representation for classification of network fragments with the targets being their parent networks. The graph image representation is based on 2D image embeddings of adjacency matrices. We use this image representation in two modes. First, as the input to a machine learning algorithm. Second, as the input to a pure transfer learner. Our conclusions from several datasets are that
\begin{itemize}
\item deep learning using our structured image features performs the best compared to benchmark graph kernel and classical features based methods; and,
\item pure transfer learning works effectively with minimum interference from the user and is robust against small data.
\end{itemize}
\end{abstract}

%
%



\maketitle

\section{Introduction}
\label{intro}
\newcommand{\remove}[1]{}

\begin{figure}[H]
  \centering
\psscalebox{0.475 0.475} 
{
\begin{pspicture}(8,8)
\psframe[linecolor=black, linewidth=0.04, dimen=outer](-1.5,8)(9.5,-0.25)

\rput(4,4){\includegraphics[scale=0.2]{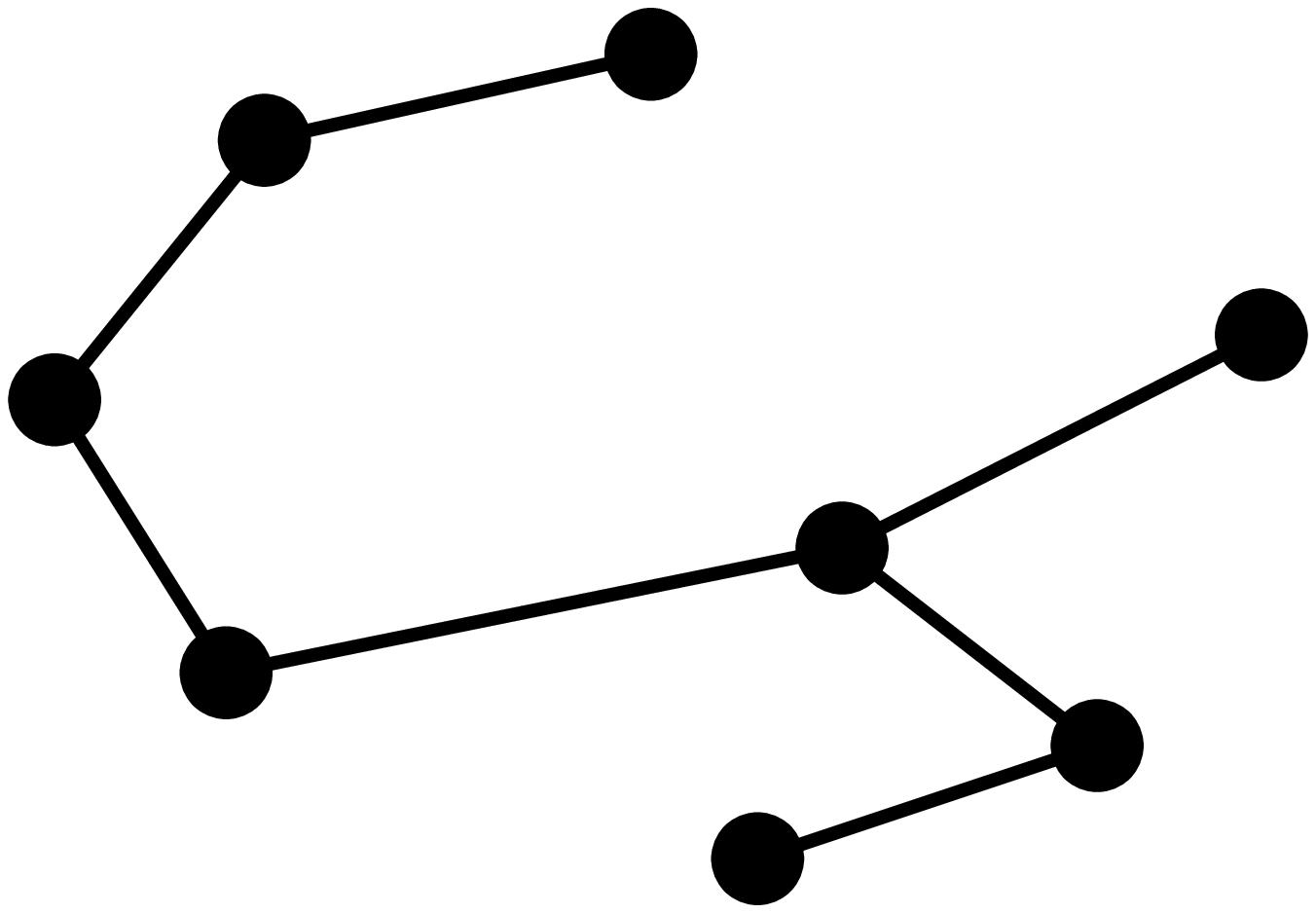}}
\rput(4,4){\fontsize{100}{100}\selectfont \textcolor{red}{?}}

\psline[linecolor=black, linewidth=0.1, arrowsize=0.05291667cm 2.0,arrowlength=1.4,arrowinset=0.0]{->}(5.53,5.29) (6.30,5.93)
\rput(7.06,6.57){\LARGE Citation}

\psline[linecolor=black, linewidth=0.1, arrowsize=0.05291667cm 2.0,arrowlength=1.4,arrowinset=0.0]{->}(4.35,5.97) (4.52,6.95)
\rput(4.61,7.45){\LARGE Gowalla}

\psline[linecolor=black, linewidth=0.1, arrowsize=0.05291667cm 2.0,arrowlength=1.4,arrowinset=0.0]{->}(3.00,5.73) (2.50,6.60)
\rput(2.25,7.03){\LARGE Facebook}

\psline[linecolor=black, linewidth=0.1, arrowsize=0.05291667cm 2.0,arrowlength=1.4,arrowinset=0.0]{->}(2.12,4.68) (1.18,5.03)
\rput(0.24,5.57){\LARGE Web Graph}

\psline[linecolor=black, linewidth=0.1, arrowsize=0.05291667cm 2.0,arrowlength=1.4,arrowinset=0.0]{->}(2.12,3.32) (1.18,2.97)
\rput(0,2.86){\LARGE \shortstack{Terrorist\\Network}}

\psline[linecolor=black, linewidth=0.1, arrowsize=0.05291667cm 2.0,arrowlength=1.4,arrowinset=0.0]{->}(3.00,2.27) (2.50,1.40)
\rput(2.15,0.87){\LARGE Wikipedia}

\psline[linecolor=black, linewidth=0.1, arrowsize=0.05291667cm 2.0,arrowlength=1.4,arrowinset=0.0]{->}(4.35,2.03) (4.52,1.05)
\rput(4.61,0.55){\LARGE Amazon}

\psline[linecolor=black, linewidth=0.1, arrowsize=0.05291667cm 2.0,arrowlength=1.4,arrowinset=0.0]{->}(5.53,2.71) (6.30,2.07)
\rput(7.06,1.43){\LARGE DBLP}

\psline[linecolor=black, linewidth=0.1, arrowsize=0.05291667cm 2.0,arrowlength=1.4,arrowinset=0.0]{->}(6.00,4.00) (7.00,4.00)
\rput(8.3,4){\LARGE \shortstack{Road\\Network}}

\end{pspicture}
}
\caption{The Problem}
\label{which}
\end{figure}

With the advent of big data, graphical representation of information has gained popularity. Being able to classify graphs has applications in many domains. We ask (Figure \ref{which}), ``Given a \emph{small} piece of a parent network, is it possible to identify the parent network?''

We address this problem using structured image representations of graphs.

Adjacency matrices are notoriously bad for machine learning. It is easy to see why, from the unstructured image of a small fragment of a road network, in Figure (a) below. Though the road network is structured, the random image would convey little or no information to machine learning algorithms (in the image, a black pixel at position $(i,j)$ corresponds to an edge between nodes $i$ and $j$). 

\begin{center}
{
\begin{tabular}{cc@{\hspace*{40pt}}cc}
\parbox{2in}{\vspace*{-60pt}\centering\centerline{(a)}

    Unstructured image of \\ adjacency matrix
    (road network)}
&  
\fboxsep0pt\fbox{\includegraphics[scale=0.25]{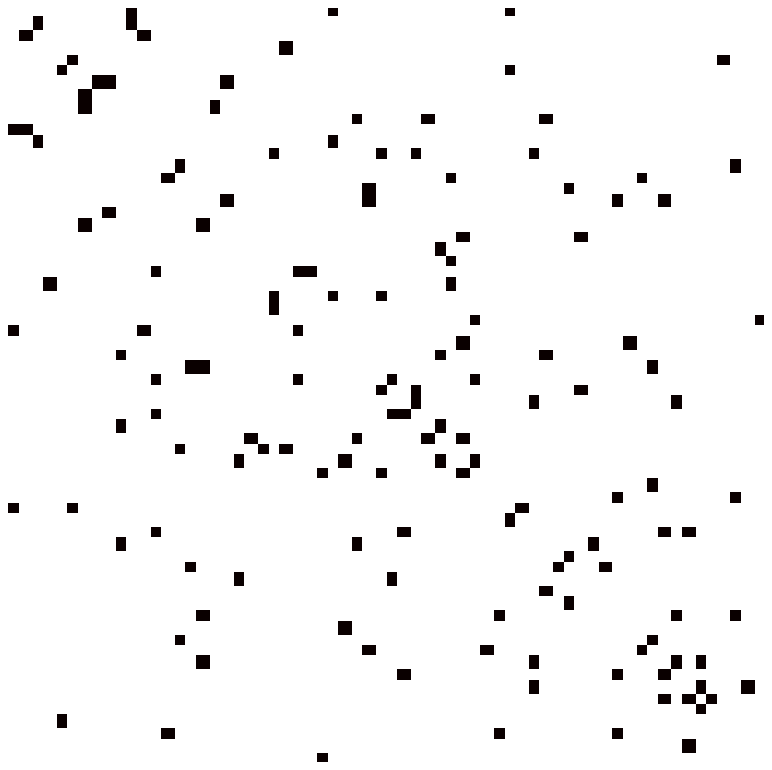}}
\\
\parbox{2in}{\vspace*{-60pt}\centering
  \centerline{(b)}

  Structured image of \\ adjacency matrix
  (road network)}
&
\fboxsep0pt\fbox{\includegraphics[scale=0.25]{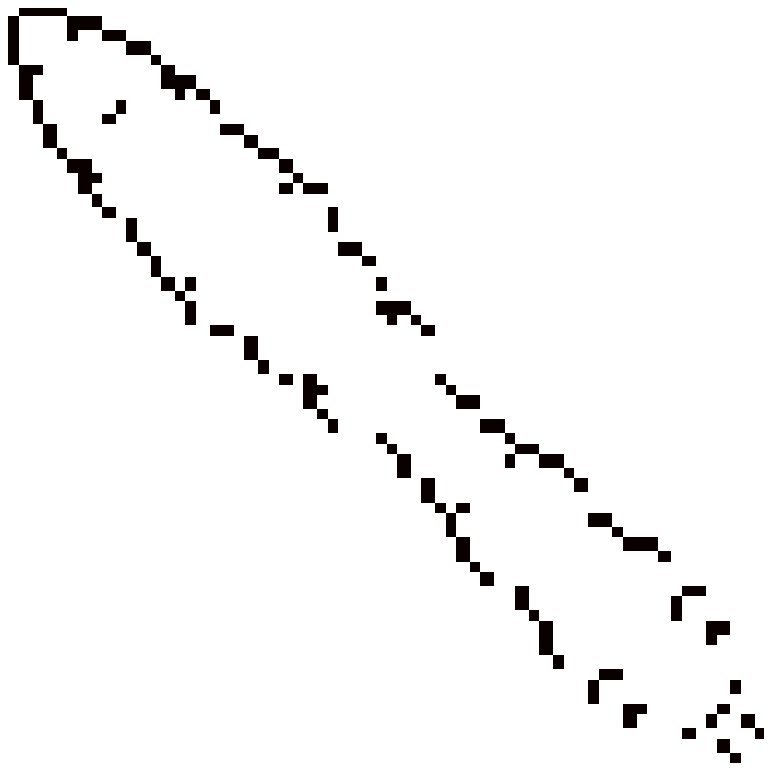}}
\end{tabular}
}
\end{center}

Reordering the vertices (Figure (b) above) gives a much more structured image \emph{for the same subgraph as in (a)}. Now, the potential to learn distinguishing properties of the subgraph is evident. We propose to exploit this very observation to solve a basic graph problem (see Figure \ref{which}). The datasets mentioned in Figure \ref{which} are discussed in Section \ref{datasets}.

\begin{figure*}
\begin{center}
\begin{subfigure}{.5\textwidth}
\begin{center}
\begin{minipage}[b]{.5\linewidth}
\vspace{2pt}
\centering
\includegraphics[width=1in]{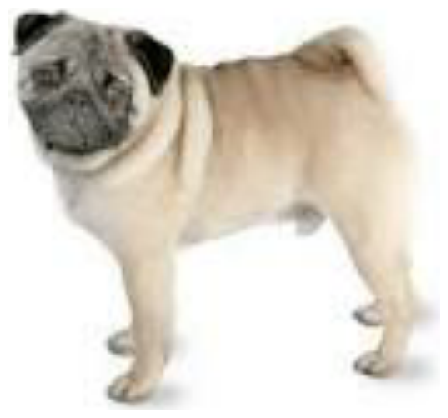}
\end{minipage}%
\begin{minipage}[b]{.5\linewidth}
\vspace{0pt}
\centering
\hspace{-1cm}
\begin{tabular}[b]{c|c}
Classification & Score\tabularnewline
\hline 
Pug & 0.75\tabularnewline
Bull Mastiff & 0.13\tabularnewline
Brabancon Giffon & 0.04\tabularnewline
French Bulldog & 0.02\tabularnewline
Muzzle & 0.01\tabularnewline
\end{tabular}
\end{minipage}
\end{center} 
\caption{An image of a dog}
\label{fig:pug-and-classification}
\end{subfigure}%
\begin{subfigure}{.5\textwidth}
\begin{center}
\begin{minipage}[b]{.4\linewidth}
\vspace{2pt}
\centering
\fboxsep0pt\fbox{\includegraphics[scale=0.28]{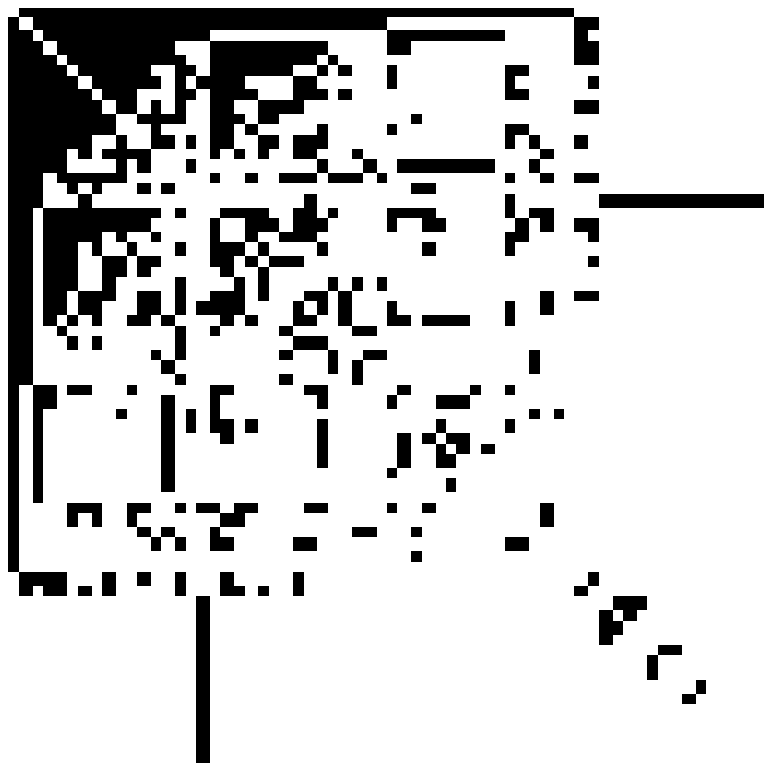}}
\end{minipage}%
\begin{minipage}[b]{.4\linewidth}
\centering
\begin{tabular}[b]{c|c}
Classification & Score\tabularnewline
\hline 
Window Screen & 0.29\tabularnewline
Digital Clock & 0.07\tabularnewline
Window Shade & 0.06\tabularnewline
Scoreboard & 0.05\tabularnewline
Oscilloscope & 0.04\tabularnewline
\end{tabular}
\end{minipage}
\end{center} 
\centering
\caption{Structured image of a Facebook subgraph}
\label{fig:embedded-graph-classification}
\end{subfigure}
\caption{Maximally specific Caffe label vectors of a dog and our structured image of a Facebook subgraph}
\label{fig:pug-and-graph}
\end{center}
\end{figure*}

We stress that both images are \emph{lossless} representations of the \emph{same} adjacency matrix. We use the structured image to classify subgraphs in two modes: \begin{inparaenum}[(i)]
\item Deep learning models on the structured image representation as input.
\item The structured image representation is used as input to a transfer learner (Caffe: see Section \ref{Caffe}) in a \emph{pure} transfer learning setting without any change to the Caffe algorithm. Caffe outputs top-$k$ categories that best describe the image. For real world images, these Caffe-descriptions are human friendly as seen in Figure \ref{fig:pug-and-classification}. However, for network-images,  Caffe gives a description which doesn't really have intuitive meaning (Figure \ref{fig:embedded-graph-classification}). We map the Caffe-descriptions to vectors. This allows us to compute similarity between network images using the similarity between Caffe description-vectors (see Section \ref{sec:methodology}).
\end{inparaenum}

\subsection{Our Contributions}
The essential difference between our work and previous approaches is that we transform graph classification into image classification. We propose an image representation of the adjacency matrix as input to machine learning algorithms for graph classification, yielding top performance. We further show that this representation is powerful enough to serve as input to a pure transfer learner that has been trained in a \emph{completely unrelated image domain}.

\textbf{The Adjacency Matrix Image Representation.} Given a sample subgraph from a parent network, the first step is to construct the image representation. We illustrate the workflow below.

\vspace{0.3cm}
\psscalebox{1 1} 
{
\hspace{-0.4cm}
\begin{pspicture}(14,3)

\rput(1.3,2){\includegraphics[scale=0.15]{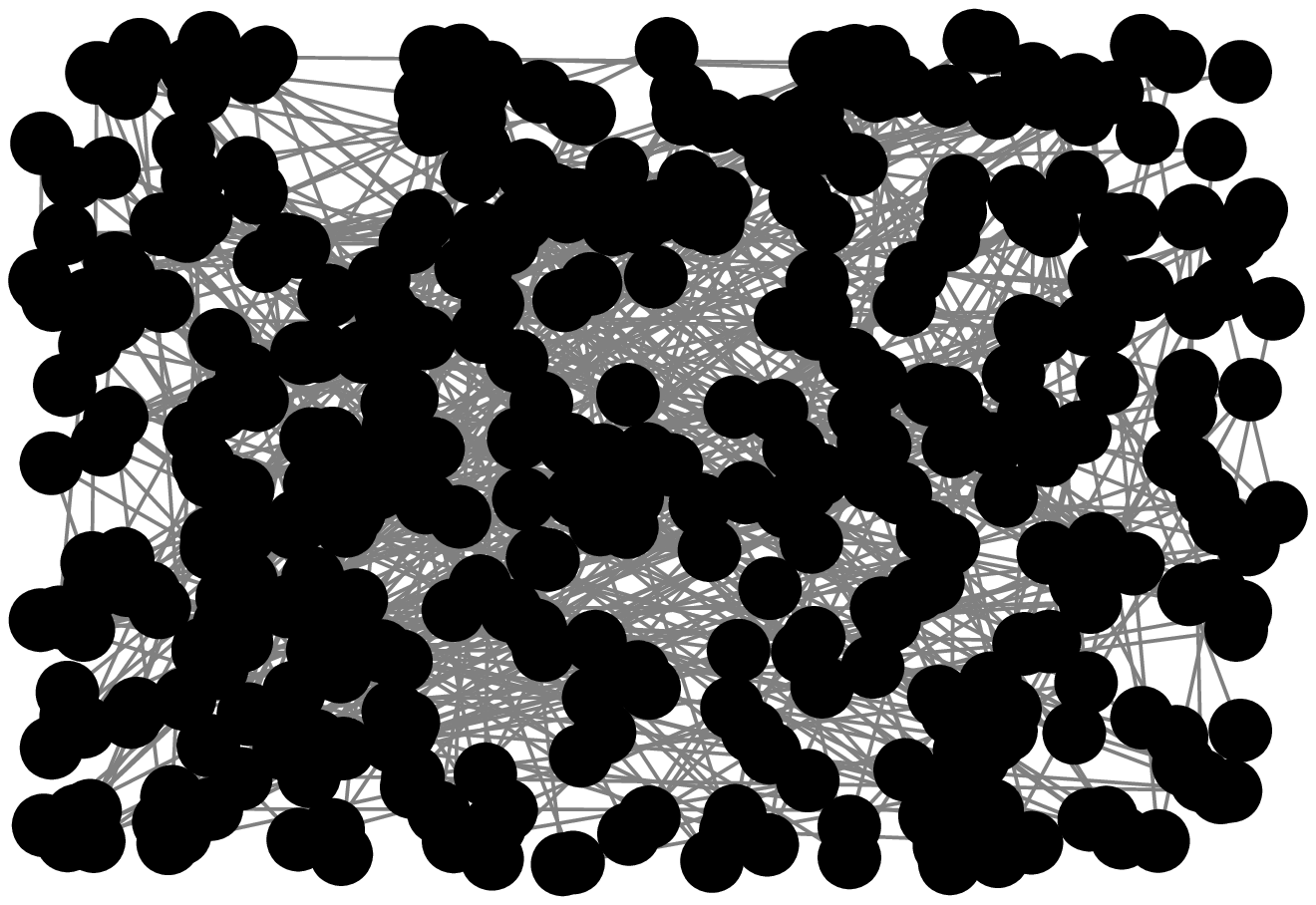}}
\rput(1.3,3){Parent Network}

\pscircle[linecolor=red, linewidth=0.08, linestyle=dotted, dimen=outer](1.3,2){0.35}
\psline[linecolor=red, linewidth=0.04, linestyle=dotted, arrowsize=0.05291667cm 2.0,arrowlength=1.4,arrowinset=0.0]{->}(1.8,2)(4.75,2)

\rput(6,2){\includegraphics[scale=0.15]{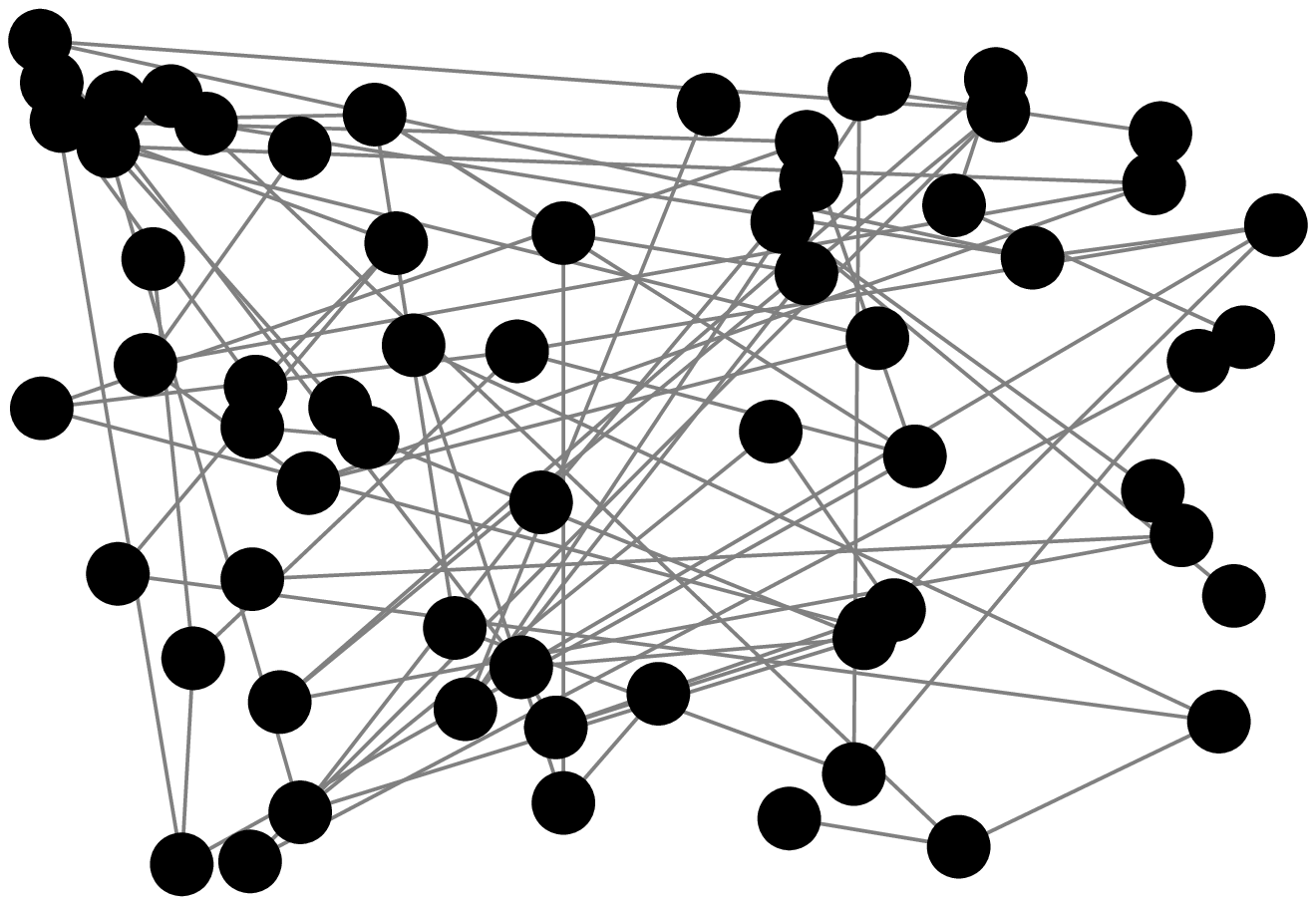}}
\rput(6,3){Small Subgraph}

\psline[linecolor=black, linewidth=0.04, arrowsize=0.05291667cm 2.0,arrowlength=1.4,arrowinset=0.0]{->}(7,2)(8,2)(8,-0.5)(7,-0.5)
\rput{90}(7.75,0.75){Image}
\rput{90}(8.25,0.75){Embedding}

\rput(6,-0.5){\fboxsep0pt\fbox{\includegraphics[scale=0.25]{road_unsorted}}}
\rput(6,-2){Raw Image}

\psline[linecolor=black, linewidth=0.04, arrowsize=0.05291667cm 2.0,arrowlength=1.4,arrowinset=0.0]{->}(4.75,-0.5)(2.5,-0.5)
\rput(3.75,-1){Image}
\rput(3.75, -1.25){Structuring}

\rput(1.3,-0.5){\fboxsep0pt\fbox{\includegraphics[scale=0.25]{road_sorted}}}
\rput(1.3,-2){Structured Image}
\end{pspicture}
}
\vspace{1.7cm}

We use the novel method proposed in~\cite{wu2016network} which produces an adjacency matrix that is invariant to permutations of the vertices in the adjacency matrix. The image is simply a ``picture'' of this permutation-invariant adjacency matrix.

\textbf{Deep Learning Using Adjacency Matrix Images.} We train deep image classifiers (see Section \ref{classifiers}) on our structured image representation as shown below. We compared several methods, including graph kernel classifiers and classifiers based on standard topological features. Our image representation performs best.

  \psscalebox{1 1} 
{
\hspace{0.2cm}
\begin{pspicture}(13,1.6)
\rput(0.7,0){\fboxsep0pt\fbox{\includegraphics[scale=0.3]{road_sorted}}}

\psline[linecolor=black, linewidth=0.04, arrowsize=0.05291667cm 2.0,arrowlength=1.4,arrowinset=0.0]{->}(2.2,1)(3.5,1)

\psframe[linecolor=black, linewidth=0.04, dimen=outer](4,0.2)(7.4,1.2)
\rput(5.7,0.7){\shortstack[l]{Network Signatures\\from Deep Learning}}

\psline[linecolor=black, linewidth=0.04, arrowsize=0.02291667cm 2.0,arrowlength=1.4,arrowinset=0.0]{->}(5.7,0.2)(5.7,-0.2)

\psframe[linecolor=black, linewidth=0.04, dimen=outer](4,-1.2)(7.4,-0.2)
\rput(5.7,-0.7){\shortstack[l]{Classification}}
\end{pspicture}
}
\vspace{1.1cm}

\textbf{Transfer Learning Using Adjacency Matrix Images.} When data is scarce or there are many missing labels, a popular option is transfer learning to leverage knowledge from some other domain. Typically the other domain is closely related to the target application. It is unusual for learning in a completely unrelated domain to be transferable to a new target domain. We show that our image representation is powerful enough that one can \emph{directly transfer learn} from the real world image domain to the network domain (two completely unrelated domains). That is, our image representation provides a link between these two domains enabling classification in the graph domain to leverage the wealth of techniques available to the image domain.

The image domain has mature pre-trained models based on massive data. For example, the open-source Caffe deep learning framework is a convolutional neural network trained on the ImageNet data which can recognize everyday objects like chairs, cats, dogs etc. (\cite{krizhevsky2012imagenet}). We use Caffe \emph{as is}. Caffe is a black box that provides a distribution over image classes which we refer to as the Caffe-classification vector. The Caffe results are then mapped back into the source domain using a distance-based heuristic e.g. Jaccard distance and $K$-nearest neighbors as shown below.

  \psscalebox{1 1} 
{
\hspace{0.2cm}
\begin{pspicture}(13,1.6)
\rput(0.7,0){\fboxsep0pt\fbox{\includegraphics[scale=0.3]{road_sorted}}}

\psline[linecolor=black, linewidth=0.04, arrowsize=0.05291667cm 2.0,arrowlength=1.4,arrowinset=0.0]{->}(2.2,1)(3.5,1)

\psframe[linecolor=black, linewidth=0.04, dimen=outer](4,0.2)(7.4,1.2)
\rput(5.7,0.7){\shortstack[l]{Caffe Classification\\Vectors}}

\psline[linecolor=black, linewidth=0.04, arrowsize=0.02291667cm 2.0,arrowlength=1.4,arrowinset=0.0]{->}(5.7,0.2)(5.7,-0.2)

\psframe[linecolor=black, linewidth=0.04, dimen=outer](4,-1.2)(7.4,-0.2)
\rput(5.7,-0.7){\shortstack[l]{Classification\\(Jaccard plus $K$-NN)}}
\end{pspicture}
}
\vspace{1.3cm}

Images, \emph{not graphs},  are passed through the Caffe deep neural network, and as we shall show, one can get good performance from as little as 10\% of the training data used in the \emph{ab initio} machine learning approach. It is quite stunning that such little training data together with un-tweaked transfer learning from a completely unrelated domain can perform so well. The reason is that our image representation provides very structured images (human-recognizable) for real world networks. Though these images are not traditional images like those used in training Caffe, Caffe still maps the different structured images to different distributions over its known classes, hence we are able to \emph{transfer} this knowledge from Caffe to graph classification.

\subsection{Related Work}
\label{relwork}
\cite{wu2016network} introduced the problem we study: Can one identify the parent network from a small subgraph? How much does local information reveal about the parent graph globally? Our approach is from the supervised setting and the unsupervised transfer learning setting.

There is previous work on similar problems using graph kernels, \cite{kernelsKashima2002,marginalizedKashima2003,boostKudo2004}, which use kernels to compute similarities between graphs and then algorithms like SVM for classification. Choosing kernels is not straightforward and is certainly not ``one-size-fits-all''. Further, these kernel methods do not scale well for very large graphs. We compare with one such method proposed by \cite{shervashidze2011weisfeiler}.

Some approaches construct features from topological attributes of the subgraph (\cite{yenerLi2011}). Topological properties of social networks have been extensively studied by \cite{topologicalAhn2007,topologicalEmilio2012,topologicalSridharan2011}. The challenges are that it is difficult to come up with a ``master set'' of features that can be used to represent graphs from different domains. For example, assortativity could be an important feature in social networks while being of little significance in road networks. It is hard to identify beforehand what features need to be computed for a given problem, thus leading to a trial and error scenario. Nevertheless, we still compare with classical features trained with logistic regression.

Transfer learning is useful when a classification task in one domain can leverage knowledge learned in a related domain (see \cite{pan2010survey} for an extensive survey). \cite{raina2007self} introduced a method called \emph{self-taught learning} which takes advantage of irrelevant unlabeled data to boost performance. \cite{zhu2011heterogeneous} discuss heterogeneous transfer learning where they use information from text data to improve image classification performance. \cite{quattoni2008transfer} create a new representation from kernel distances to large unlabeled data points before performing image classification using a small subset of reference prototypes.

\paragraph{Paper Organization.}
In Section \ref{sec:methodology} we give more details of our approaches to subgraph classification. Section \ref{res} present the results comparing the performance of our
approach with other approaches. We conclude in Section \ref{cfw} with some possible future directions.

\section{Methodology}
\label{sec:methodology}

Given fragments of a large network we first obtain their structured images to construct the training examples. In the supervised setting, we learn the final classifier from this training. In the transfer setting,  we use the Caffe framework as a black-box to obtain the label-vectors. To classify a test subgraph, we first obtain its label-vector through Caffe, compute the distance between the test label-vector and the training label-vectors, and classify using majority among the nearest$-k$ training examples. We now give the details.

\subsection{Image embeddings of graphs}
\label{sec:Image}

An adjacency matrix of a graph can be thought of as a monochrome image with 1s corresponding to dark pixels and 0s corresponding to white pixels. This observation, however, is of limited practical use since it is unstructured and not permutation invariant. We suitably reorder the vertices in the graph so that structural information can be represented by spatial organization within the image. We utilize a novel technique for producing a permutation-invariant ordering of the adjacency matrix, first given in~\cite{wu2016network}. The authors describe several ways to sort an adjacency matrix like page rank, degree based sorting, etc. They show that the BFS-like approach works best. The ordering starts with the node of highest degree (ties are broken using $k$-neighborhood size for $k=2$, then $k=3$, $\dots$). Subsequently, the ordering proceeds based on a combination of shortest paths and degrees. Details can be found in~\cite{wu2016network}. The ordering scheme results in permutation-invariant adjacency matrices from which we obtain structured images.

We observe that the image embeddings have enough structure so that even the human eye can distinguish between different networks without much effort. Neural networks are highly successful in recognizing real world objects that have high level structural similarity. For example, all dogs have similar features although they are individually different. Similarly, the image embeddings of all road network subgraphs ``look" similar but are microscopically different. From a neural network's perspective, the structured image embeddings are like any other images. The fact that these structured image embeddings were obtained from adjacency matrices has no methodological impact on the neural network model.

\subsection{Network Signatures from Deep Learning}
\label{classifiers}

Our primary objective is to use deep learning to learn to classify subgraphs into their parent networks using our image representation of the subgraph. However, we also tested a wide variety of other methods as well. We describe them briefly below. 

\textbf{Deep Belief Network (DBN)}. Our DBNs (see: \cite{dbnHinton2009}) have
multiple layers of unsupervised Restricted Boltzmann Machines (RBMs),
\cite{hinton2006fast},
trained greedily and fine-tuned using back-propagation.

Typically, DBNs have an input layer, hidden layer(s) and a final output layer. For an $8 \times 8$ image, the input layer has 64 nodes. The hidden layers consist of RBMs where the output of each of RBMs are used as input to the next. Finally, the output layer contains a node for each class. The probabilities of each class label is returned. The one with the highest probability is chosen as the overall classification for the given input.

\textbf{Convolutional Neural Network (CNN)}. CNNs \cite{cnnkeras} have proven to be very effective in real world image classification tasks. The building blocks of a CNN are convolution layers, non-linear layers such as Rectified Linear Units (ReLU), pooling layers and fully connected layers for classification.  

\textbf{Stacked De-Noising Auto-Encoder (SdA)}. We used the stacked de-noising auto-encoder (SdA) based on greedy training presented in \cite{wu2016node}. In a regular multi-layer deep neural network, each layer is trained to "reconstruct" the input from the previous layer. Then, the system is fine tuned by using back-propagation. In SdA, instead of the original input, a noisy input is fed to the system.

\textbf{Diffusion-Convolutional Neural Network (DCNN)}. DCNNs introduced by \cite{dcnnAtwood2016}, work on the graphs themselves rather than the image embeddings of their adjacency matrices.  DCNNs provide a flexible tool for of graphical data that encodes node features, edge features, and purely structural information with little preprocessing. DCNNs learn diffusion-based representations from graph-structured data using a diffusion-convolution operation.

\textbf{Graph Kernels (GK)}. We use the graph kernel in \cite{shervashidze2011weisfeiler} which extracts features based on the Weisfeiler-Lehman test of isomorphism on graphs. It maps the original graph to a sequence of graphs, whose node attributes capture topological and label information. We use the implementations from \cite{graphkernels}.

\textbf{Classical Features and Logistic Regression (LR)}. We compute 15 classic features on subgraphs and train a logistic regression model \cite{sklearnlogreg} on these features. Our features are: transitivity, average clustering co-efficient, average node connectivity, edge connectivity, average eccentricity, diameter, average shortest path, average degree, fraction of single-degree nodes, average closeness centrality, central points, density, average neighbor degree and top two eigen values of the adjacency matrix.


All methods were tested in a standard supervised learning framework where $n$ training examples $(x_i,y_i)$ are given (inputs $x_i$ and targets $y_i$) DBN, CNN and SdA, the input is our image representation of the subgraph. For DCNN and GK, the input is the graph itself. For LR, the input is a vector of 15 classical features.

\subsection{Caffe-Classification and Transfer Learning}
\label{Caffe}

Caffe \cite{jia2014caffe} is a large scale project developing a deep learning framework for image classification. It has been extensively used in image classification and filter visualization, learning handwritten digits and style recognition, etc. \cite{reddit}. We use a pre-trained model that is trained on a crowd-sourced labeled data set ImageNet. As of 2016, ImageNet had more than 10 million hand-annotated images. The massive volume combined with a deep convolutional neural network gives us fine-grained discriminatory power for images. 

Given an image, Caffe gives a vector of (label, label-probability) tuples.
An example of the output for a real image is shown in Figure~\ref{fig:pug-and-classification}. Although Caffe has not been trained on image embeddings of graphs, such images nonetheless produce vectors that have sufficient discriminatory information that we extract using the post processing step (see Section \ref{postproc}). An example of a vector corresponding to a Facebook subgraph is shown in Figure~\ref{fig:embedded-graph-classification}. 

Caffe provides either \emph{maximally accurate} or \emph{maximally specific} classification. We use the \emph{maximally specific} option. Further, while we have shown cardinality-5 vectors for brevity in Figure \ref{fig:pug-and-graph}, we use cardinality-10 vectors in our experiments.

Finally, Caffe is trained on color images, while our images are
black and white (each pixel represents presence or absence of an edge).
Embeddings that encode additional information (e.g. edge weights) as RGB
values may allow us to leverage the color capability of Caffe, which
is a possible avenue for future research.

\subsubsection{Post-processing for Graph Classification}
\label{postproc}

Caffe provides a set of label vectors $L_i$ for each training network $x_i$. Each labelvector is a tuple of (label, label-probability pairs) as deemed by Caffe.
In this work, we ignore the probabilities, and treat each vector as an unordered list of labels (strings). Each training vector also has a \emph{ground truth} parent label.

We use Jaccard similarity to compute a similarity metric between two label-vectors $L_j$ and $L_k$:
\[
d(L_j, L_k) = \frac{|L_j \cap L_k|}{|L_j \cup L_k|}
\]

We leave for future work the use of more sophisticated metrics which could use the probabilities from Caffe - our goal is to demonstrate the potential of even this
simplest possible approach.

For a test graph, we get the label-vector $T$ from Caffe and then compute the $k$ nearest training vectors using the Jaccard distance $d(L_i, T)$ to each training vector $L_i$ and classify using the majority class $C$ among these $k$ nearest training vectors (ties are broken randomly). The test example is correctly classified if and only if its ground truth matches $C$. One advantage of the $k$ nearest neighbor approach is that it seamlessly handles multiclass problems.

\subsection{Data}
\label{datasets}

\begin{figure*}
\begin{subfigure}{.5\textwidth}
  \centering

\begin{adjustbox}{minipage=\linewidth,scale=1}
\begin{subfigure}{.33\textwidth}
  \centering
  \includegraphics[width=.8\linewidth]{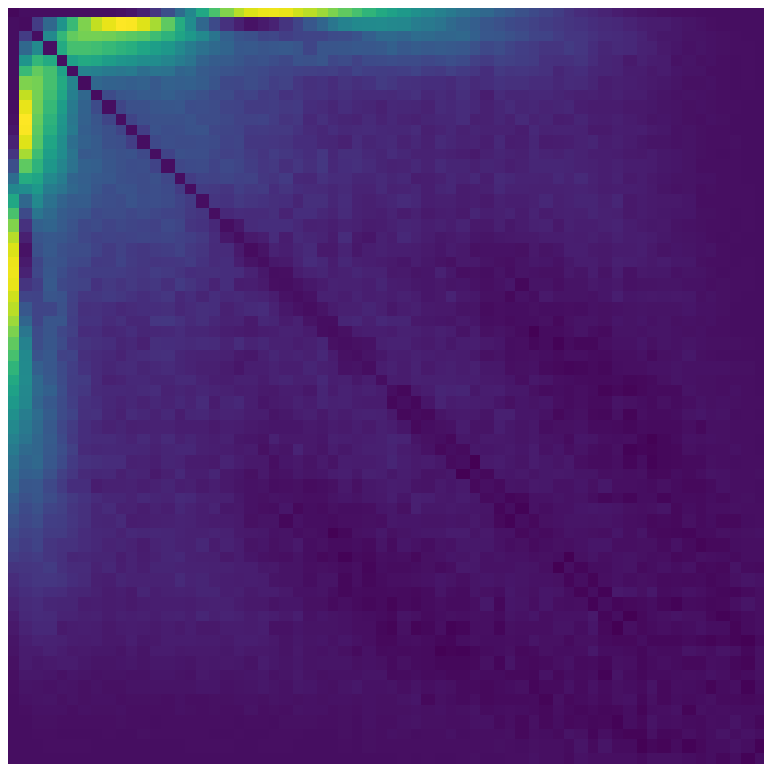}
  \caption{ }
  \label{eigcitation}
\end{subfigure}%
\begin{subfigure}{.33\textwidth}
  \centering
  \includegraphics[width=.8\linewidth]{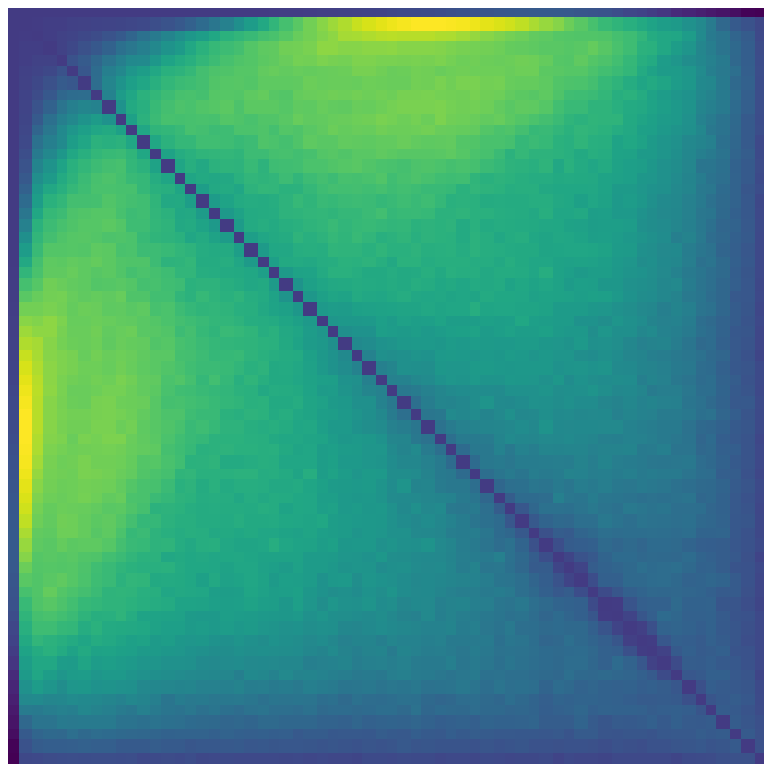}
  \caption{ }
  \label{eigfacebook}
\end{subfigure}%
\begin{subfigure}{.33\textwidth}
  \centering
  \includegraphics[width=.8\linewidth]{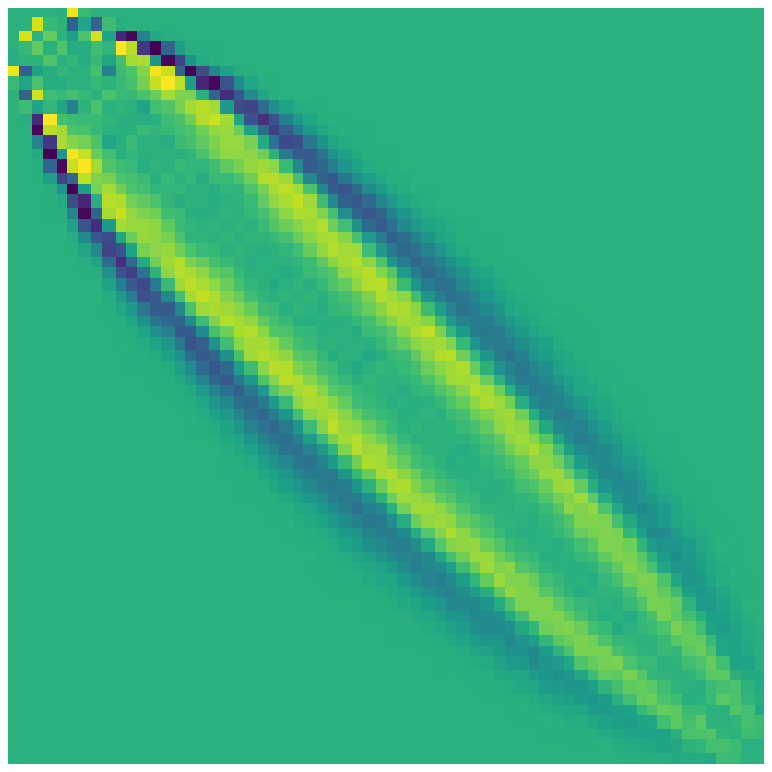}
  \caption{ }
  \label{eigroad}
\end{subfigure}

\begin{subfigure}{.33\textwidth}
  \centering
  \includegraphics[width=.8\linewidth]{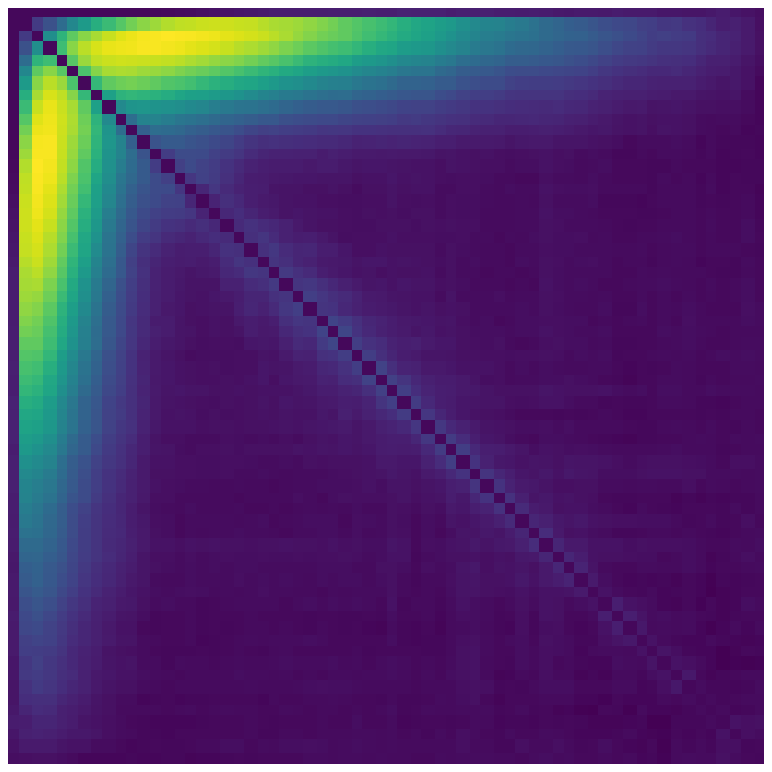}
  \caption{ }
  \label{eigweb}
\end{subfigure}%
\begin{subfigure}{.33\textwidth}
  \centering
  \includegraphics[width=.8\linewidth]{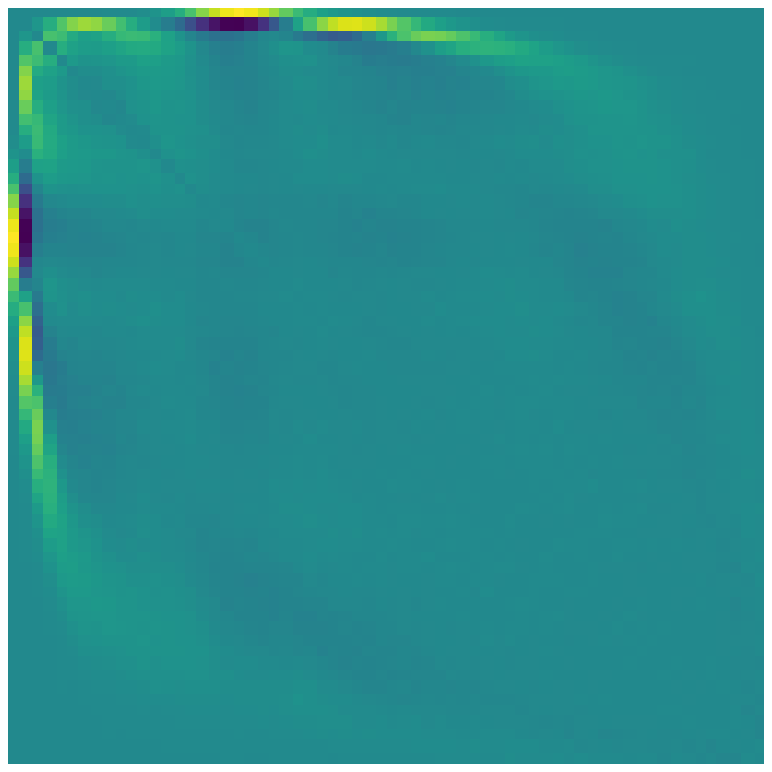}
  \caption{ }
  \label{eigwiki}
\end{subfigure}%
\begin{subfigure}{.33\textwidth}
  \centering
  \includegraphics[width=.8\linewidth]{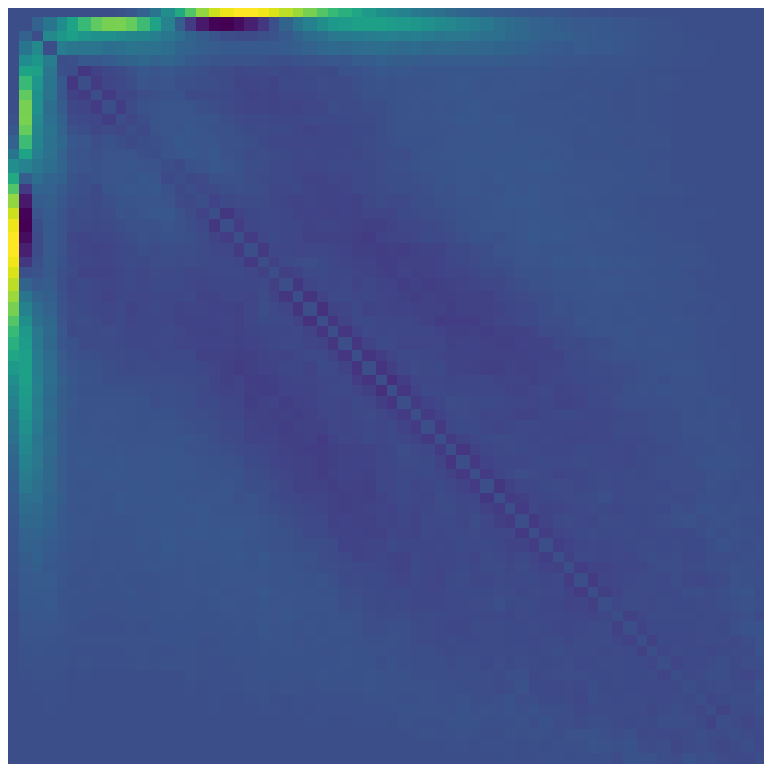}
  \caption{ }
  \label{eigamazon}
\end{subfigure}

\begin{subfigure}{.33\textwidth}
  \centering
  \includegraphics[width=.8\linewidth]{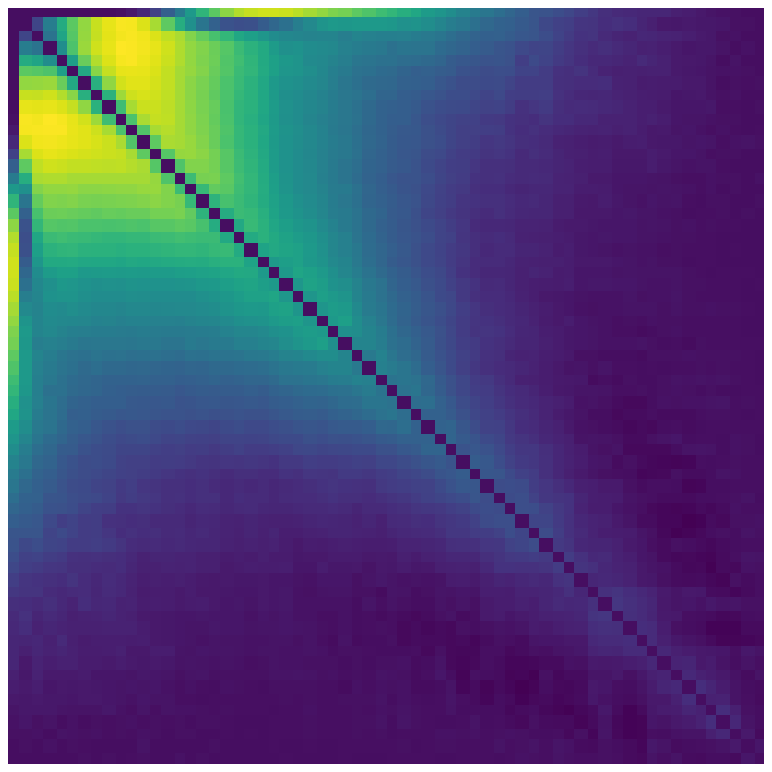}
  \caption{ }
  \label{eigdblp}
\end{subfigure}%
\begin{subfigure}{.33\textwidth}
  \centering
  \includegraphics[width=.8\linewidth]{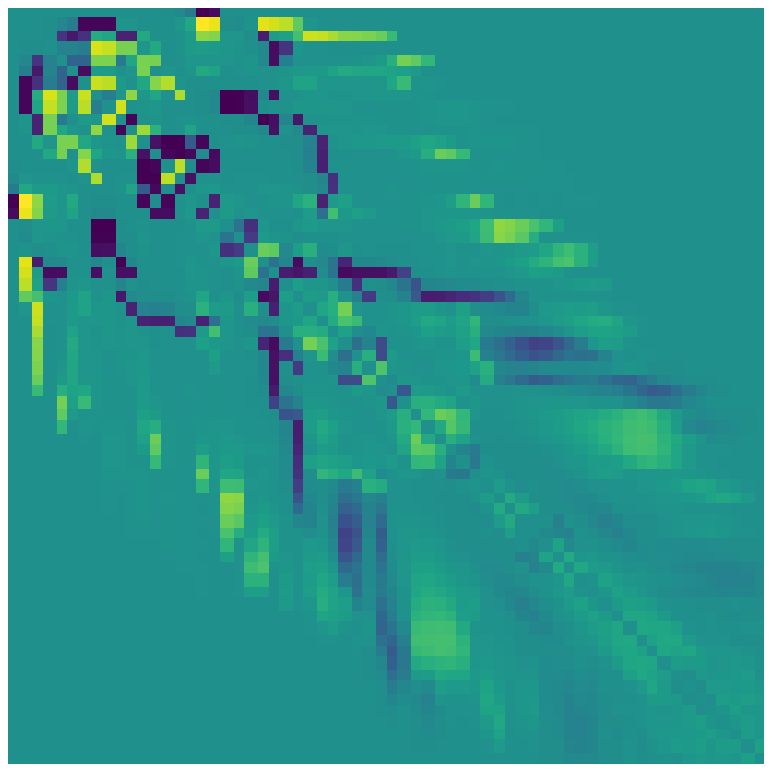}
  \caption{ }
  \label{eigaq}
\end{subfigure}%
\begin{subfigure}{.33\textwidth}
  \centering
  \includegraphics[width=.8\linewidth]{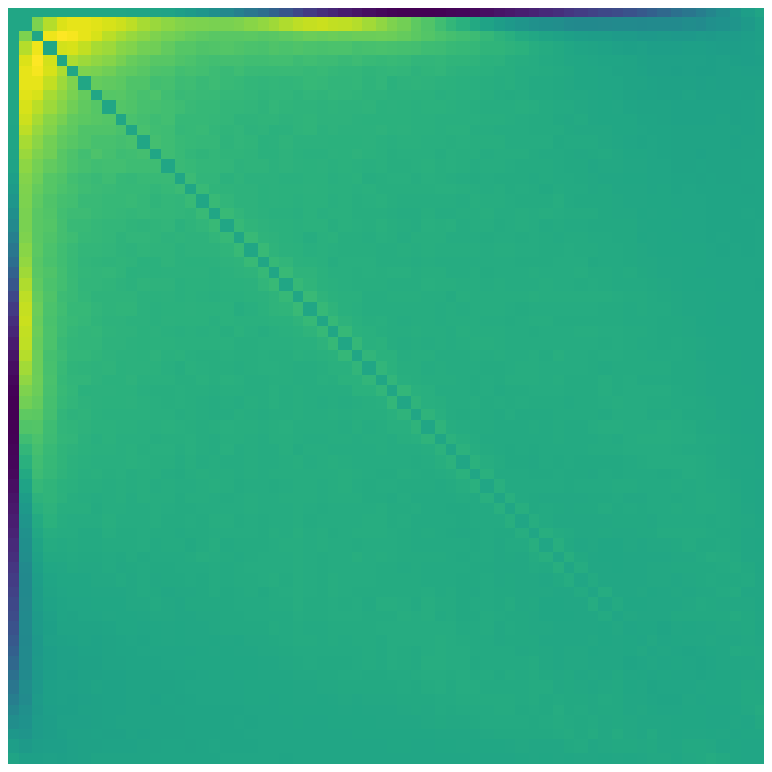}
  \caption{ }
  \label{eiggo}
\end{subfigure}
\end{adjustbox}
\setcounter{subfigure}{0}%
  \caption{Top principal component}
  \label{eig}
\end{subfigure}%
\begin{subfigure}{.5\textwidth}
  \centering
\setcounter{subfigure}{0}%
\begin{adjustbox}{minipage=\linewidth,scale=1}
\begin{subfigure}{.33\textwidth}
  \centering
  \fboxsep0pt\fbox{\includegraphics[width=.8\linewidth]{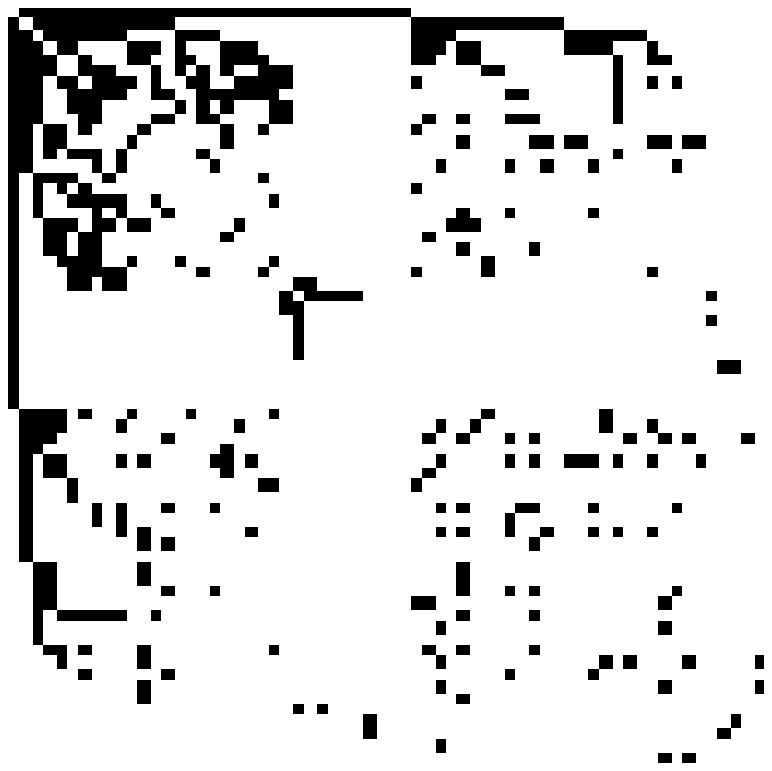}}
  \caption{ }
  \label{sortcitation}
\end{subfigure}%
\begin{subfigure}{.33\textwidth}
  \centering
  \fboxsep0pt\fbox{\includegraphics[width=.8\linewidth]{facebook_sorted}}
  \caption{ }
  \label{sortfacebook}
\end{subfigure}%
\begin{subfigure}{.33\textwidth}
  \centering
  \fboxsep0pt\fbox{\includegraphics[width=.8\linewidth]{road_sorted}}
  \caption{ }
  \label{sortroad}
\end{subfigure}

\begin{subfigure}{.33\textwidth}
  \centering
  \fboxsep0pt\fbox{\includegraphics[width=.8\linewidth]{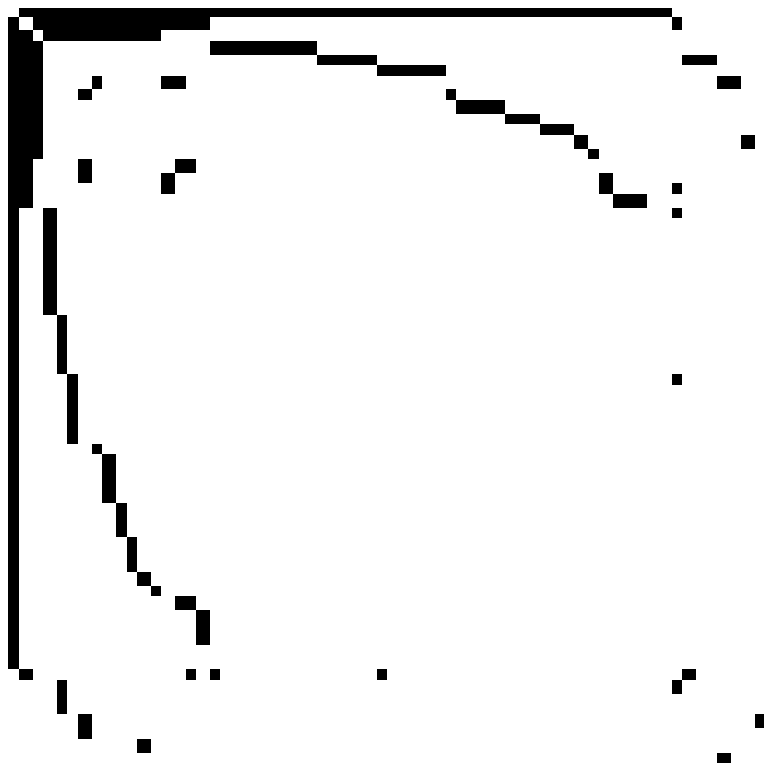}}
  \caption{ }
  \label{sortweb}
\end{subfigure}%
\begin{subfigure}{.33\textwidth}
  \centering
  \fboxsep0pt\fbox{\includegraphics[width=.8\linewidth]{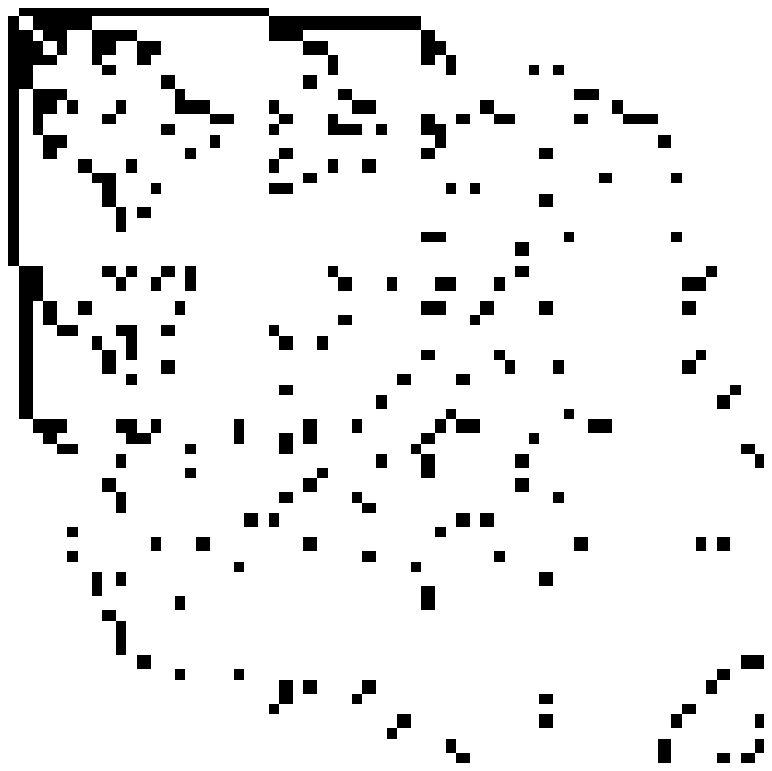}}
  \caption{ }
  \label{sortwiki}
\end{subfigure}%
\begin{subfigure}{.33\textwidth}
  \centering
  \fboxsep0pt\fbox{\includegraphics[width=.8\linewidth]{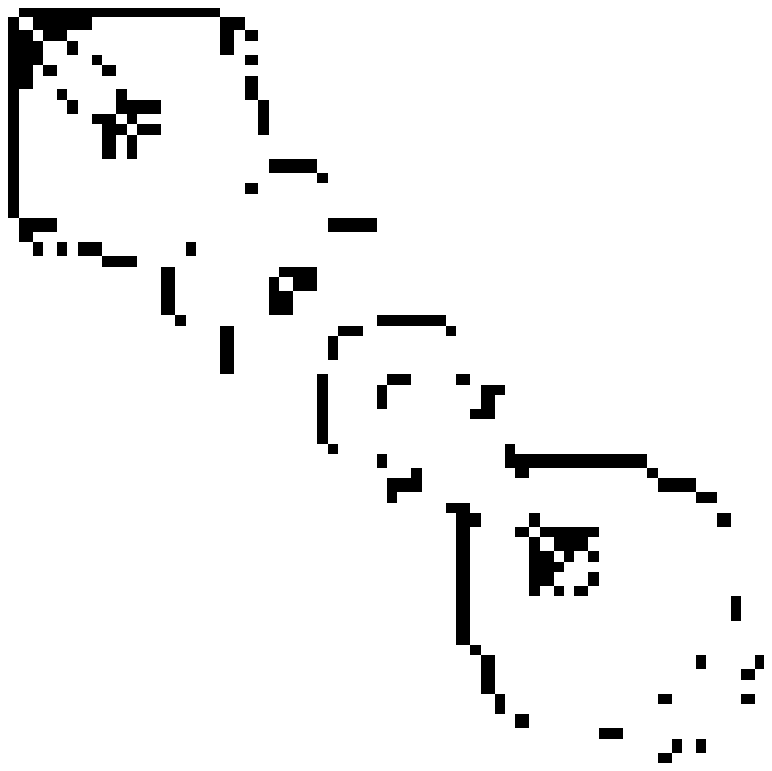}}
  \caption{ }
  \label{sortamazon}
\end{subfigure}

\begin{subfigure}{.33\textwidth}
  \centering
  \fboxsep0pt\fbox{\includegraphics[width=.8\linewidth]{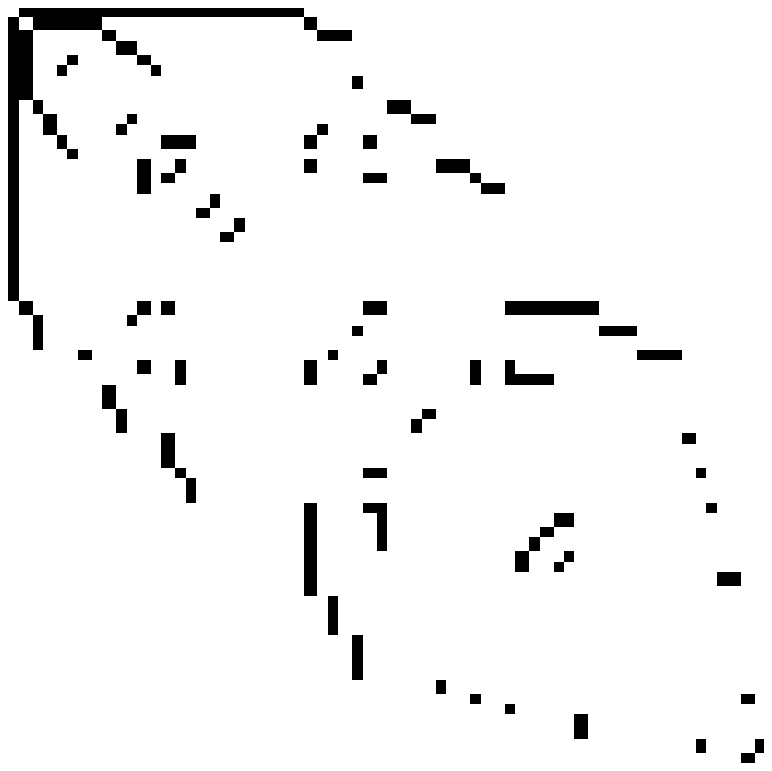}}
  \caption{ }
  \label{sortdblp}
\end{subfigure}%
\begin{subfigure}{.33\textwidth}
  \centering
  \fboxsep0pt\fbox{\includegraphics[width=.8\linewidth]{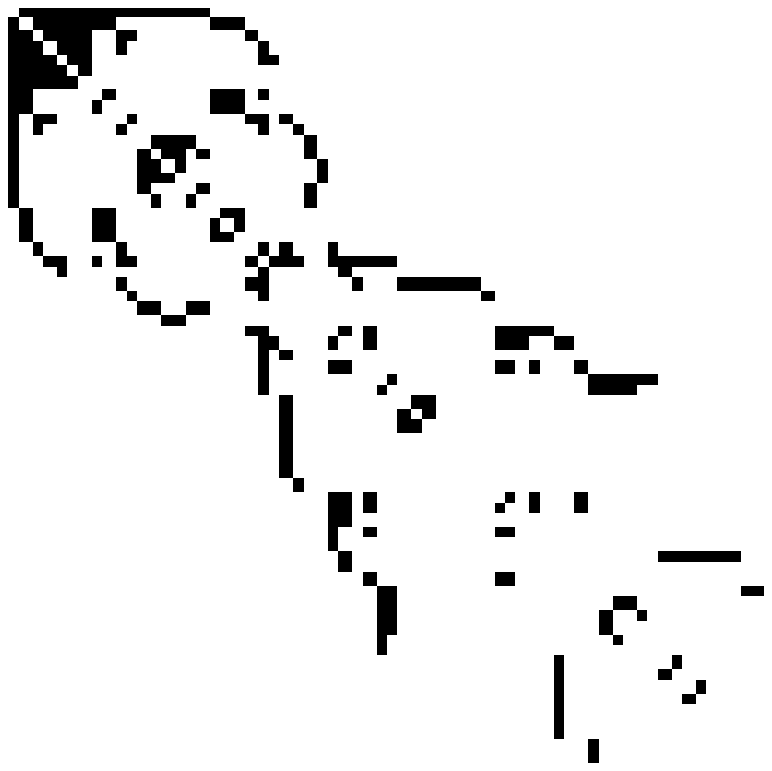}}
  \caption{ }
  \label{sortaq}
\end{subfigure}%
\begin{subfigure}{.33\textwidth}
  \centering
  \fboxsep0pt\fbox{\includegraphics[width=.8\linewidth]{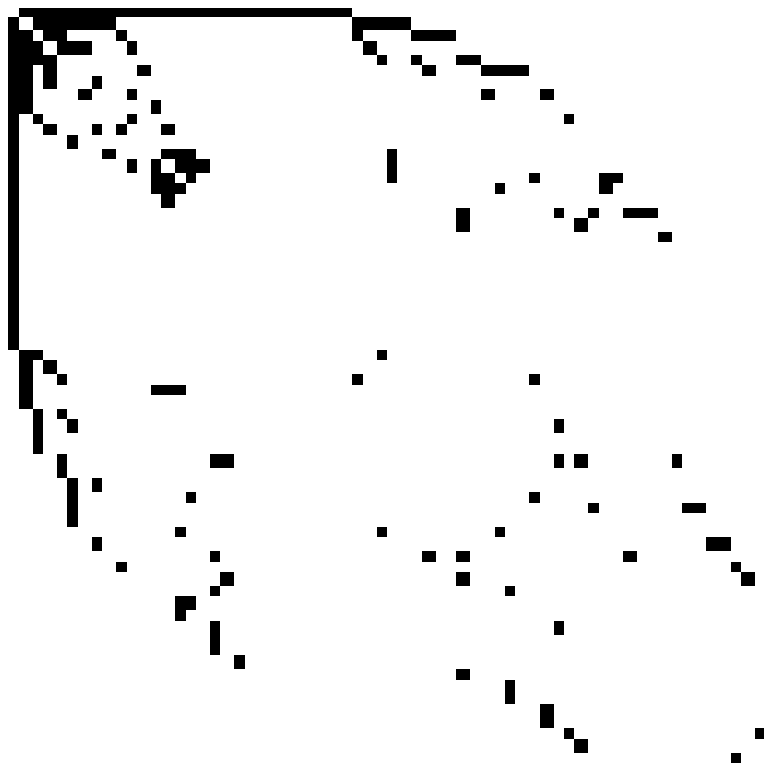}}
  \caption{ }
  \label{sortgo}
\end{subfigure}
\end{adjustbox}
\setcounter{subfigure}{1}%
  \caption{Structured images of sample subgraphs}
  \label{struct}
\end{subfigure}
\caption{Structured images and the top principal component. (a) Citation; (b) Facebook; (c) Road Network; (d) Web; (e) Wikipedia;
(f) Amazon; (g) DBLP; (h) Terrorist Network; (i) Gowalla}
\label{eigvstruct}
\end{figure*}

We used a variety of parent graphs, from citation networks to social networks to e-commerce networks (see Table \ref{datasetstable} and the Appendix). Our classifiers learn network-signatures from the image representations. To gain insight into these network signatures, we show the top principal component
from each network class in Figure \ref{eig}. To get these principal components, we used a PCA analysis of the vectorized images in each class (similar to \cite{turk1991face}).

In Figure \ref{struct}, we show the structured images of specific 64-node subgraphs from each class. These are adjacency matrices that have gone through the structuring process described in Section \ref{sec:Image}. Observe that the images for different classes are both well structured and quite different from each other. This is why learning is able to perform well at classifying the subgraphs, see in Section~\ref{res}.

\begin{table}
\centering
\begin{tabular}{c|c|c|l}
Dataset & \# Nodes & \# Edges & Reference(s)\tabularnewline
\hline 
Citation & 34,546 & 421,578 & \cite{citationLeskovec2005,citationGehrke2003}\tabularnewline
Facebook & 4039 & 88,234 & \cite{facebookMcAuley2012}\tabularnewline
Road Network & 1,088,092 & 1,541,898 & \cite{roadLeskovec2009}\tabularnewline
Web & 875,713 & 5,105,039 & \cite{roadLeskovec2009}\tabularnewline
Wikipedia & 4,604 & 119,882 & \cite{wikiWest2012,wiki2West2009}\tabularnewline
Amazon & 334,863 & 925,872 & \cite{amazonLeskovec2007}\tabularnewline
DBLP & 317,080 & 1,049,866 & \cite{dblpYang2012}\tabularnewline
Terrorist Net. & 271 & 756 & \cite{jjatt}\tabularnewline
Gowalla & 196,591 & 950,327 & \cite{cho2011friendship}\tabularnewline
\end{tabular}
\caption{Datasets}
\label{datasetstable}
\end{table}

\section{Experimental Setup and Results}
\label{res}

We now describe our experimental setup and present the results from the supervised and transfer approaches.

\subsection{Deep Supervised Image Classifiers}
\label{supresults}

We perform graph classification for the above mentioned parent networks. We perform a random walk on each of these networks $5,000$ times until we get the desired number of nodes for a training subgraph, denoted by $n$. We tried $n \in \{8$, $16$, $32, 64\}$. So, with $9$ networks and $5,000$ samples per network, we create $4$ datasets with $45,000$ samples each. Each dataset is of the size $45,000 \times n \times n$.

For a given dataset, we randomly chose $\frac{1}{3}$ of it for training, validation and testing respectively. The accuracy score is defined as the ratio of the sum of the principal diagonal entries of the confusion matrix over the sum of all the entries of the confusion matrix (sometimes called the error matrix \cite{stehman1997selecting}). We report the accuracy score for each classifier for $n = 64$ in the following table. The full confusion matrices are in the Appendix.

\begin{center}
\begin{tabular}{c|cccccc}
 & CNN & SdA & DBN & DCNN & GK & LR\tabularnewline
\hline 
Acc. & 0.86 & 0.79 & 0.81 & 0.45 & 0.73 & 0.83\tabularnewline
\end{tabular}
\end{center}

\begin{figure}[h]
\centering
  \includegraphics[scale=0.47]{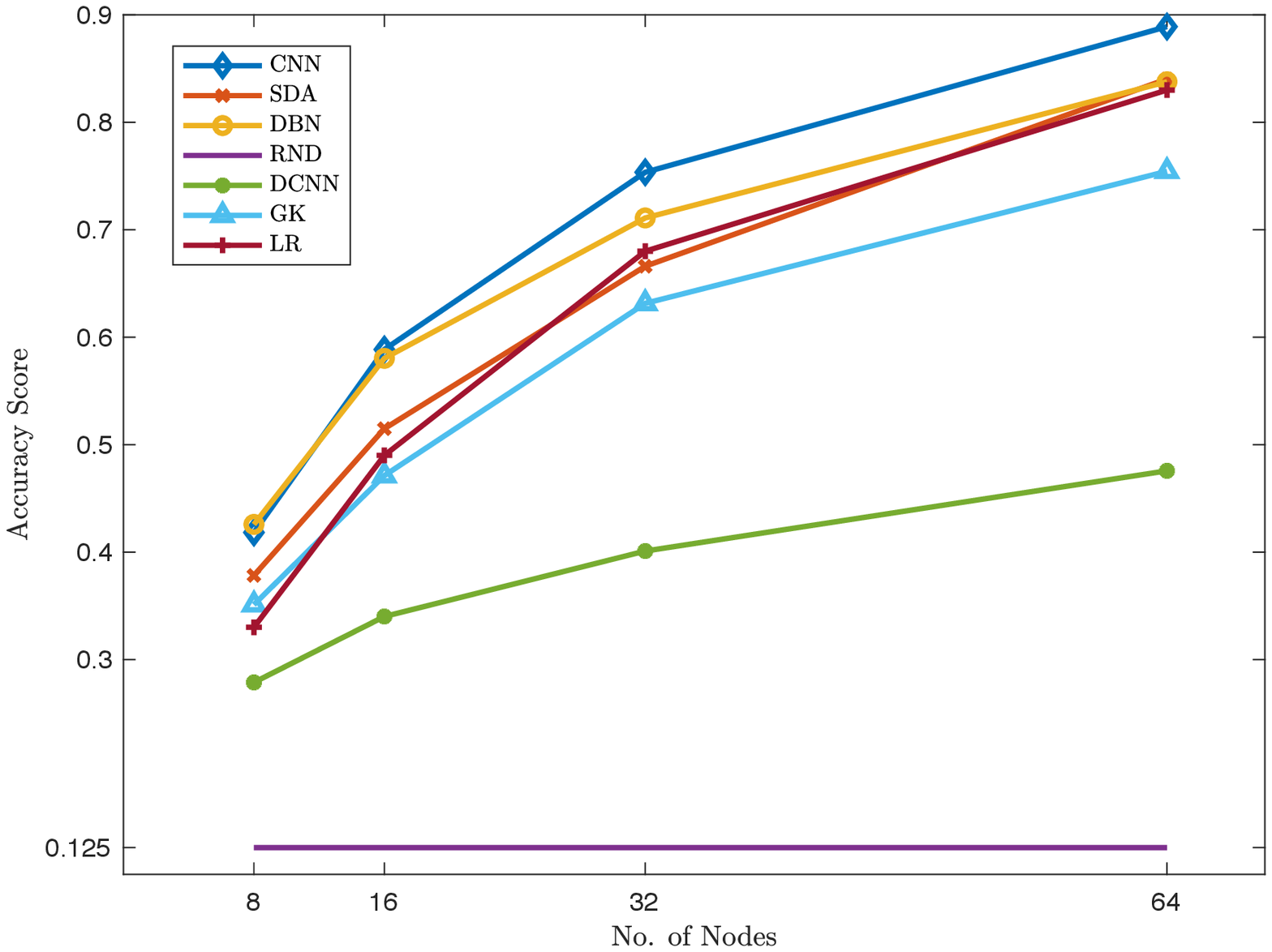}
  \caption{Classification accuracy increases as $n$ increases}
  \label{summarypic}
\end{figure}

CNN was the best performing classifier while DCNN and GK\footnote{It uses \texttt{libsvm}'s \cite{chang2011libsvm} ``one-vs-one" method for multiclass classification} were the poorest performers. Off-the-shelf deep learning with our graph image features is the top performer. Figure \ref{summarypic} summarizes the performance of all the methods. As expected, we obtained higher accuracy as the subgraph size increases, because the subgraph contains more information (the flat line in Figure \ref{summarypic} is for random guessing). We point out that even with $8-$node subgraphs do significantly better than random.

We also note that although LR performs well, it is very hard to choose the features when graphs come from different domains. One set of features that worked best in one scenario may not be the best in another. So, off-the-shelf CNN, DBN or SdA with our image features are an easy choice to make as they require little effort and deliver top performance. We also considered hybrid training sets with subgraphs of different sizes. We observed that the performance was better when the mixture had more examples with higher $n$. This is inline with the results in Figure \ref{summarypic}. Interested readers can refer to the Appendix for more details.

Graph kernel and feature-based methods performed better than DCNN, but not as well as the image embedding based methods. Kernel methods are complex and usually slow and the fact that they did not perform spectacularly do not make them attractive. One may notice that the accuracy scores for LR are very comparable to SdA. However, we would like to remark that LR is a \emph{lossy} method since it approximates the graph and boils it down to a handful of features. It is very hard to decide which features must be used and the choice may vary for graphs in different domains in order to get optimal results. However, our approach is completely \emph{lossless}. The structured image representation of the graph has every bit of the information the adjacency matrix does. So, we would not have to make any compromises to get the best out of the data. 

Out of the 4 neural network classification models, DCNN is the only one that takes a graph as an input instead of images. In fact, it takes two inputs: an adjacency matrix and a \emph{design} matrix. Design matrix takes information (average degree, clustering co-efficient etc.) about each node in the adjacency matrix. We observed no change in performance when we did not provide any information in the design matrix. This is because these \emph{accessory} properties can be calculated from the graph which is already an input. The neural network is expected to have \emph{learned} these features already.

\subsection{Transfer Learning}
\label{unsupresults}

We present our experimental results for transfer learning. We show that our transfer learning approach is highly resilient to sparse training data. We achieve a respectable accuracy even when only 10\% of the data was used for training.

We treat Caffe as a black-box that requires for transfer learning. Transfer learning helps because when one does not have access to ample data to train classifiers from scratch.

We did the following experiments: (a) differentiating networks of similar theme/type (i. Terrorist Net. vs Facebook, ii. Citation vs DBLP and iii. Web vs Wiki); and (b) full multiclass classification. Tables \ref{table:aq-vs-facebook}, \ref{table:citations-vs-dblp}, and \ref{table:web-vs-wiki} show the results of classification between similarly-themed networks. We see varying degrees of success: Wiki vs Web gives a very high accuracy of 95\%. The other two are respectable as well, with 90\% and 80\% accuracies.


\begin{table}
\centering
\begin{subtable}{.35\textwidth}
\centering
\begin{tabular}{l|rrr}
Data & Prc. & Rec. & F1\tabularnewline
\hline 
Ter. Net. & 0.84 & 0.99 & 0.91\tabularnewline
FB & 0.99 & 0.81 & 0.89\tabularnewline
\hline 
\textbf{Acc.} & \multicolumn{3}{c}{\textbf{90.3\%}}\tabularnewline
\end{tabular}%
\caption{Terrorist Net. vs Facebook}
\label{table:aq-vs-facebook}
\end{subtable}

\begin{subtable}{.35\textwidth}
\centering
\begin{tabular}{l|rrr}
Data & Prc. & Rec. & F1\tabularnewline
\hline 
Cit. & 0.90 & 0.67 & 0.76\tabularnewline
DBLP & 0.74 & 0.92 & 0.82\tabularnewline
\hline 
\textbf{Acc.} & \multicolumn{3}{c}{\textbf{79.51\%}}\tabularnewline
\end{tabular}%
\caption{Citations vs DBLP}
\label{table:citations-vs-dblp}
\end{subtable}

\begin{subtable}{.35\textwidth}
\centering
\begin{tabular}{l|rrr}
Data & Prc. & Rec. & F1\tabularnewline
\hline 
Wiki & 0.96 & 0.93 & 0.94\tabularnewline
Web & 0.93 & 0.97 & 0.94\tabularnewline
\hline 
\textbf{Acc.} & \multicolumn{3}{c}{\textbf{94.57\%}}\tabularnewline
\end{tabular}
\caption{Wiki vs Web}
\label{table:web-vs-wiki}
\end{subtable}
\caption{Pairwise classification for transfer learning}
\end{table}

Table \ref{table:all-six} shows that transfer learning does not do quite as well when it comes to multiclass classification, which is challenging in general. Our particular approach of using the majority rule may be somewhat more impacted since now the correct class has to outnumber several other classes.

\begin{table}
\centering
\begin{center}
\begin{tabular}{l|rrr}
Data & Prc. & Rec. & F1\tabularnewline
\hline 
Wiki & 0.71 & 0.79 & 0.75\tabularnewline
Web & 0.66 & 0.73 & 0.69\tabularnewline
DBLP & 0.57 & 0.59 & 0.58\tabularnewline
Terrorist Net. & 0.48 & 0.70 & 0.57\tabularnewline
Citations & 0.36 & 0.18 & 0.24\tabularnewline
Facebook & 0.84 & 0.66 & 0.74\tabularnewline
\hline 
\textbf{Accuracy} & \multicolumn{3}{c}{\textbf{61\%}}\tabularnewline
\end{tabular}
\end{center}
\caption{Multiclass classification for transfer learning}
\label{table:all-six}
\end{table}

\begin{figure}
  \centering
  \includegraphics[scale=0.7]{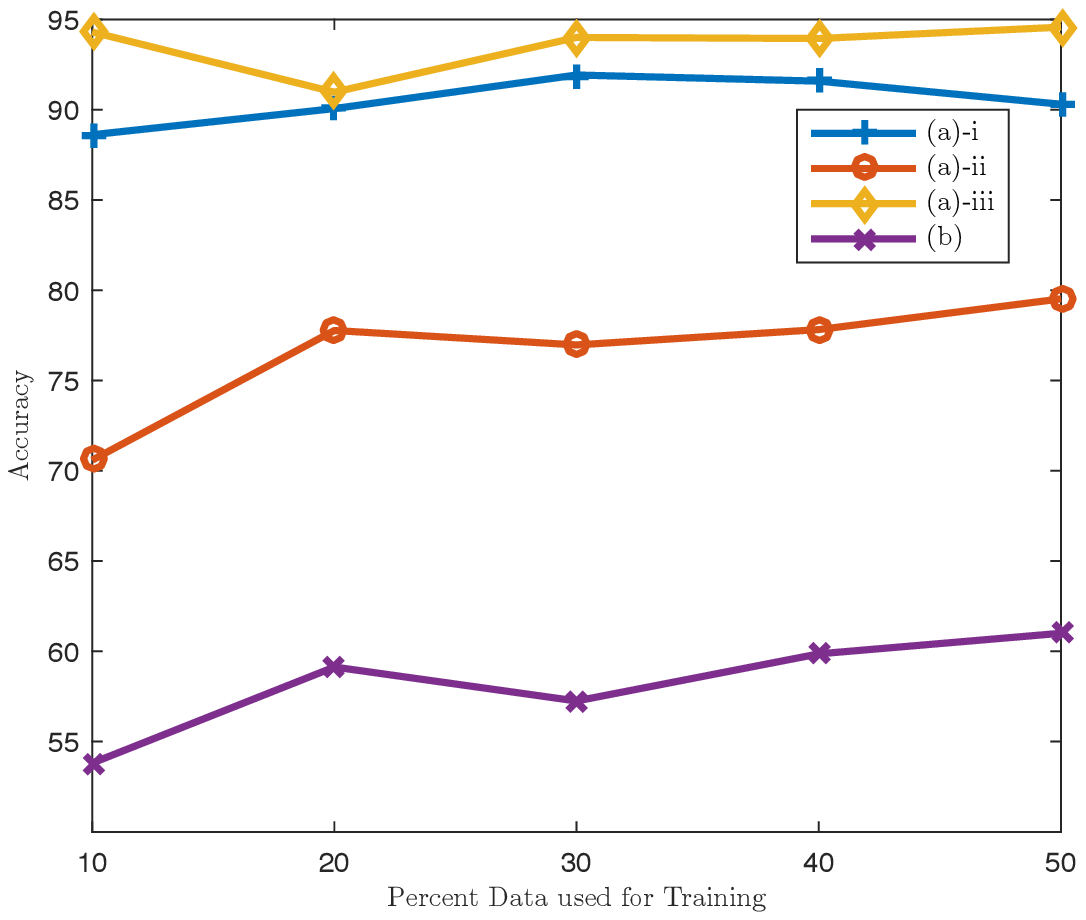}
  \caption{Accuracy of transfer learning. (a)-i: Terrorist Net. vs Facebook, (a)-ii: Citation vs DBLP and (a)-iii: Web vs Wiki and (b): multi-way classification.}
\label{fig:acc-vs-training}
\end{figure}

Figure~\ref{fig:acc-vs-training} plots the accuracy numbers pairwise classification as we progressively increase the proportion of data used for identifying k-nearest neighbors (the rest of the  data is used for testing). The takeaway message is that transfer learning works, and is robust to small training data. Even 10\% of the data gives good performance. Most learning techniques, especially deep neural networks are sensitive to training data size. The relative insensitivity of transfer learning is because it leverages a \emph{pre-trained} recognition engine in the image domain which has already been trained with a massive dataset. Hence transfer learning approach is very robust and resilient to sparse training data.

Finally, we study the impact of $k$, the neighborhood size for the majority rule. The details are in the Appendix, but the take away fact is that as long as $k > 15$, the performance is robust to $k$.

\section{Conclusion and Future Work}
\label{cfw}
Our experiments overwhelmingly show that our structured image representation of graphs achieves successful graph classification with ease. The image representation is \emph{lossless}, that is the image embeddings contain all the information in the corresponding adjacency matrix. Our results also show that even with very small subgraphs, deep network models are able to extract network signatures from our highly structured images. Specifically, with just 64-node subgraphs from networks with up to 1 million nodes, we were able to predict the parent network with $> 90\%$ accuracy. Even with 8-node subgraphs, accuracies are significantly better than random. Further, we demonstrated that the image embedding approach provides many advantages over graph kernel and feature-based methods.

We also presented an approach to graph classification using transfer learning from a completely different domain. Our approach converts graphs into 2D image embeddings and uses a pre-trained image classifier (Caffe) to obtain label-vectors. In a range of experiments with real-world data sets, we have obtained accuracies from 70\% to 94\% for pairwise classification and 61\% for multiclass classification. Further, our approach is highly resilient to training-to-test ratio, that is, can work with little training data. Our results show that such an approach is very promising, especially for applications where training data is not readily available (e.g. terrorist networks).

Future work includes improvements to the transfer learning by improving the distance function between label-vectors, as well as using the probabilities from Caffe. Further, we would also look to generalize this approach to other domains, for example classifying radio frequency map samples using transfer learning.

\textbf{Acknowledgments.} This research was supported by the Army Research Laboratory under Cooperative Agreement W911NF-09-2-0053 (the ARL-NSCTA). The views and conclusions contained in this document are those of the authors and should not be interpreted as representing the official policies, either expressed or implied, of the Army Research Laboratory or the U.S. Government. The U.S. Government is authorized to reproduce and distribute reprints for government purposes notwithstanding any copyright notation here on.

\bibliographystyle{acmart}
\bibliography{acmart}

\appendix
\section{Appendix}
\label{appendix}

\subsection{Deep Supervised Image Classifiers Results Contd.}
We present the confusion matrices of all classifiers for 64-node subgraphs in Table \ref{confmats}. A confusion matrix $C$ is such that $C_{i, j}$ is equal to the number of observations known to be in class $i$ but predicted (\emph{confused}) to be in class $j$.

\subsection{Hybrid Datasets}
\begin{figure}[h!]
\centering
\includegraphics[scale=0.6]{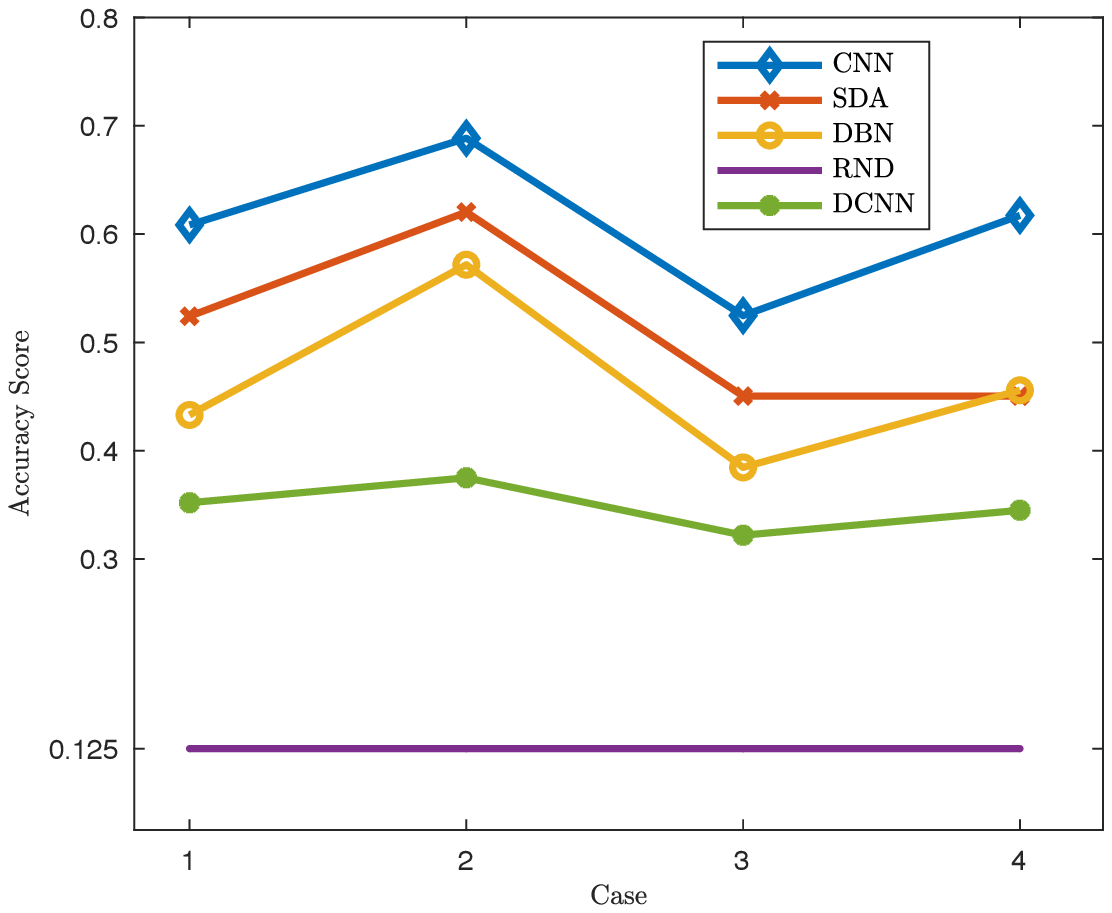}
\caption{Performance on hybrid datasets}
\label{combopic}
\end{figure}

We present the detailed results for different mixtures of datasets that we experimented with in the supervised setting. Although the performance deteriorates when different $n$'s are mixed, the relative ordering of the methods w.r.t. their performances remains the same. Table below shows the 4 cases and as Figure \ref{combopic} shows, classification accuracy increases as the contribution by subgraphs with higher $n$ increases.

\begin{center}
\begin{tabular}{c|cccc}
Nodes & Case 1 & Case 2 & Case 3 & Case 4\tabularnewline
\hline
8 & 25\% & 10\% & 40\% & 10\%\tabularnewline
16 & 25\% & 20\% & 30\% & 40\%\tabularnewline
32 & 25\% & 30\% & 20\% & 40\%\tabularnewline
64 & 25\% & 40\% & 10\% & 10\%\tabularnewline
\end{tabular}
\end{center}

\subsection{Transfer Learning Results Contd.}
\begin{figure}[h!]
  \centering
  \includegraphics[scale=0.6]{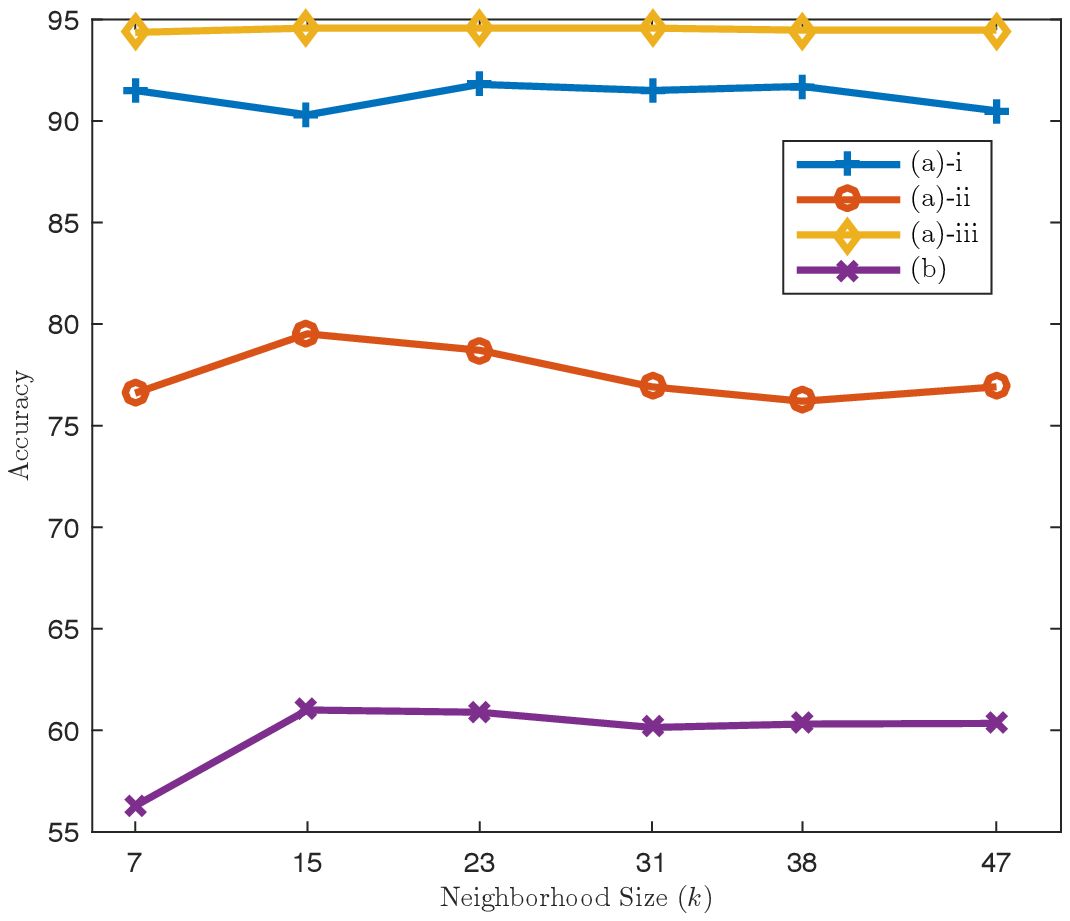}
  \caption{Accuracy vs neighborhood size ($k$)}
  \label{fig:acc-vs-k}
\end{figure}

Figure~\ref{fig:acc-vs-k} shows the variation of accuracy results when $k$ is varied in steps of 8 from $k=7$ to $k=47$, with the base case $k=15$ used for the tabulated results above included for comparison. As can be seen, except for a couple of cases with $k=7$ providing a somewhat lower accuracy, the variation is within 1\% of the base case. Thus, as long as $k > 15$, we do not have to worry too much about tuning $k$, showing that the approach is robust.

\subsection{Datasets}

\textbf{Citation}. This citation network is from Arxiv HEP-PH (high energy physics phenomenology). If a paper $i$ cites paper $j$, the graph contains a directed edge from $i$ to $j$. There are $34,546$ nodes with $421,578$ edges. See: \cite{citationLeskovec2005,citationGehrke2003}

\textbf{Facebook}. This social network contains "friends lists" from Facebook. There are $4039$ people (nodes) and there is an undirected edge between nodes if they are friends. There are $88,234$ such edges. See: \cite{facebookMcAuley2012}

\textbf{Road Network}. This is a road network of Pennsylvania. Intersections and endpoints are represented by nodes, and the roads connecting these intersections are represented by undirected edges. There are $1,088,092$ nodes and $1,541,898$ edges in this network. See: \cite{roadLeskovec2009}

\textbf{Web}. Nodes in this network represent web pages and directed edges represent hyperlinks between them. There are $875,713$ nodes and $5,105,039$ edges. See: \cite{roadLeskovec2009}

\textbf{Wikipedia}. This is a Wikipedia hyperlink graph. A condensed version of Wikipedia was used in the collection of this dataset. There are $4,604$ articles (nodes) with $119,882$ links (edges) between them. See: \cite{wikiWest2012,wiki2West2009}

\textbf{Amazon}. This is a product co-purchase network of amazon.com. The nodes are products sold on amazon.com. There is an undirected edge between two products if they are frequently co-purchased. There are $334,863$ nodes and $925,872$ edges. See: \cite{amazonLeskovec2007}

\textbf{DBLP}. This is a co-authorship network. It has authors for its nodes and there is an undirected edge between them if they have co-authored at least one paper. There are $317,080$ nodes and $1,049,866$ edges. See: \cite{dblpYang2012}

\textbf{Terrorist Network}. We use the ``Al Qaeda Operations Attack Series 1993-2003, Worldwide" dataset consisting of $271$ nodes (participants of Al Qaeda terrorist group) with $756$ links between them. See: \cite{jjatt}

\textbf{Gowalla}. Gowalla was a location-based social networking website where users shared their locations by checking-in. The friendship network is undirected and consists of $196,591$ nodes and $950,327$ edges. See: \cite{cho2011friendship}

\begin{table*}
\begin{adjustbox}{minipage=\textwidth,scale=0.8}
\begin{subtable}[t]{\textwidth}
\hspace{1cm}
\begin{tabular}{c|ccccccccc}
 & Citation & Facebook & Road Net. & Web & Wikipedia & Amazon & DBLP & Terrorist Net. & Gowalla\tabularnewline
\hline 
Citation & \textbf{1110} & 17 & 2 & 6 & 58 & 292 & 42 & 0 & 139\tabularnewline
Facebook & 11 & \textbf{1649} & 0 & 1 & 0 & 5 & 1 & 0 & 0\tabularnewline
Road Net. & 0 & 0 & \textbf{1666} & 0 & 0 & 1 & 0 & 0 & 0\tabularnewline
Web & 39 & 8 & 0 & \textbf{1392} & 1 & 69 & 41 & 0 & 117\tabularnewline
Wikipedia & 28 & 0 & 0 & 0 & \textbf{1627} & 1 & 0 & 0 & 10\tabularnewline
Amazon & 248 & 2 & 4 & 42 & 1 & \textbf{1177} & 74 & 1 & 117\tabularnewline
DBLP & 34 & 3 & 0 & 12 & 3 & 82 & \textbf{1513} & 0 & 20\tabularnewline
Terrorist Net. & 3 & 0 & 0 & 0 & 0 & 18 & 14 & \textbf{1632} & 0\tabularnewline
Gowalla & 225 & 5 & 1 & 41 & 48 & 128 & 27 & 0 & \textbf{1192}\tabularnewline
\end{tabular}
\caption{CNN}
\end{subtable}

\begin{subtable}[t]{\textwidth}
\hspace{1cm}
\begin{tabular}{c|ccccccccc}
 & Citation & Facebook & Road Net. & Web & Wikipedia & Amazon & DBLP & Terrorist Net. & Gowalla\tabularnewline
\hline 
Citation & \textbf{804} & 28 & 0 & 64 & 135 & 369 & 90 & 18 & 158\tabularnewline
Facebook & 9 & \textbf{1625} & 0 & 24 & 0 & 1 & 7 & 1 & 0\tabularnewline
Road Net. & 0 & 0 & \textbf{1667} & 0 & 0 & 0 & 0 & 0 & 0\tabularnewline
Web & 49 & 51 & 0 & \textbf{1268} & 17 & 104 & 44 & 1 & 133\tabularnewline
Wikipedia & 40 & 0 & 0 & 1 & \textbf{1568} & 5 & 1 & 0 & 51\tabularnewline
Amazon & 223 & 0 & 0 & 69 & 15 & \textbf{997} & 169 & 13 & 180\tabularnewline
DBLP & 44 & 29 & 0 & 31 & 3 & 155 & \textbf{1364} & 8 & 33\tabularnewline
Terrorist Net. & 10 & 0 & 0 & 0 & 0 & 51 & 20 & \textbf{1586} & 0\tabularnewline
Gowalla & 199 & 7 & 0 & 124 & 129 & 170 & 42 & 1 & \textbf{995}\tabularnewline
\end{tabular}
\caption{SdA}
\end{subtable}

\begin{subtable}[t]{\textwidth}
\hspace{1cm}
\begin{tabular}{c|ccccccccc}
 & Citation & Facebook & Road Net. & Web & Wikipedia & Amazon & DBLP & Terrorist Net. & Gowalla\tabularnewline
\hline 
Citation & \textbf{1010} & 20 & 1 & 40 & 87 & 271 & 80 & 0 & 157\tabularnewline
Facebook & 9 & \textbf{1626} & 0 & 20 & 0 & 0 & 11 & 0 & 1\tabularnewline
Road Net. & 1 & 0 & \textbf{1665} & 0 & 0 & 1 & 0 & 0 & 0\tabularnewline
Web & 63 & 21 & 0 & \textbf{1353} & 13 & 75 & 32 & 0 & 110\tabularnewline
Wikipedia & 78 & 0 & 0 & 1 & \textbf{1480} & 5 & 1 & 0 & 101\tabularnewline
Amazon & 330 & 2 & 3 & 71 & 12 & \textbf{969} & 147 & 3 & 129\tabularnewline
DBLP & 54 & 8 & 0 & 27 & 6 & 127 & \textbf{1411} & 1 & 33\tabularnewline
Terrorist Net. & 13 & 0 & 0 & 0 & 0 & 26 & 27 & \textbf{1601} & 0\tabularnewline
Gowalla & 246 & 3 & 0 & 105 & 113 & 161 & 27 & 1 & \textbf{1011}\tabularnewline
\end{tabular}
\caption{DBN}
\end{subtable}

\begin{subtable}[t]{\textwidth}
\hspace{1cm}
\begin{tabular}{c|ccccccccc}
 & Citation & Facebook & Road Net. & Web & Wikipedia & Amazon & DBLP & Terrorist Net. & Gowalla\tabularnewline
\hline 
Citation & \textbf{192} & 520 & 235 & 12 & 31 & 87 & 479 & 28 & 82\tabularnewline
Facebook & 18 & \textbf{1607} & 13 & 10 & 0 & 17 & 1 & 0 & 1\tabularnewline
Road Net. & 0 & 0 & \textbf{1667} & 0 & 0 & 0 & 0 & 0 & 0\tabularnewline
Web & 37 & 297 & 29 & \textbf{1160} & 3 & 32 & 66 & 5 & 38\tabularnewline
Wikipedia & 163 & 358 & 98 & 0 & \textbf{213} & 93 & 710 & 5 & 26\tabularnewline
Amazon & 161 & 220 & 347 & 94 & 11 & \textbf{126} & 554 & 37 & 116\tabularnewline
DBLP & 47 & 282 & 286 & 2 & 31 & 20 & \textbf{891} & 34 & 74\tabularnewline
Terrorist Net. & 51 & 9 & 814 & 0 & 26 & 0 & 73 & \textbf{694} & 0\tabularnewline
Gowalla & 33 & 224 & 68 & 729 & 34 & 87 & 292 & 2 & \textbf{198}\tabularnewline
\end{tabular}
\caption{DCNN}
\end{subtable}

\begin{subtable}[t]{\textwidth}
\hspace{1cm}
\begin{tabular}{c|ccccccccc}
 & Citation & Facebook & Roadnet & Web & Wikipedia & Amazon & DBLP & Terrorist Net. & Gowalla\tabularnewline
\hline 
Citation & \textbf{759} & 0 & 0 & 0 & 0 & 907 & 0 & 0 & 0\tabularnewline
Facebook & 0 & \textbf{1600} & 0 & 67 & 0 & 0 & 0 & 0 & 0\tabularnewline
Roadnet & 0 & 0 & \textbf{1667} & 0 & 0 & 0 & 0 & 0 & 0\tabularnewline
Web & 0 & 0 & 0 & \textbf{1130} & 0 & 537 & 0 & 0 & 0\tabularnewline
Wikipedia & 0 & 0 & 0 & 0 & \textbf{1399} & 0 & 0 & 0 & 267\tabularnewline
Amazon & 849 & 0 & 0 & 0 & 0 & \textbf{817} & 0 & 0 & 0\tabularnewline
DBLP & 0 & 0 & 0 & 0 & 0 & 380 & \textbf{1287} & 0 & 0\tabularnewline
Terrorist Net. & 0 & 0 & 0 & 0 & 0 & 190 & 0 & \textbf{1477} & 0\tabularnewline
Gowalla & 790 & 0 & 0 & 0 & 0 & 0 & 0 & 0 & \textbf{877}\tabularnewline
\end{tabular}
\caption{GK}
\end{subtable}

\begin{subtable}[t]{\textwidth}
\hspace{1cm}
\begin{tabular}{c|ccccccccc}
 & Citation & Facebook & Road Net. & Web & Wikipedia & Amazon & DBLP & Terrorist Net. & Gowalla\tabularnewline
\hline 
Citation & \textbf{1095} & 26 & 1 & 35 & 54 & 288 & 44 & 8 & 115\tabularnewline
Facebook & 29 & \textbf{1613} & 0 & 20 & 0 & 4 & 1 & 0 & 0\tabularnewline
Road Net. & 0 & 0 & \textbf{1667} & 0 & 0 & 0 & 0 & 0 & 0\tabularnewline
Web & 28 & 44 & 0 & \textbf{1255} & 3 & 66 & 57 & 6 & 208\tabularnewline
Wikipedia & 5 & 0 & 0 & 0 & \textbf{1656} & 0 & 0 & 0 & 5\tabularnewline
Amazon & 214 & 3 & 1 & 87 & 3 & \textbf{1063} & 138 & 42 & 115\tabularnewline
DBLP & 18 & 21 & 1 & 17 & 1 & 78 & \textbf{1465} & 31 & 35\tabularnewline
Terrorist Net. & 0 & 0 & 0 & 0 & 0 & 10 & 43 & \textbf{1614} & 0\tabularnewline
Gowalla & 170 & 7 & 1 & 98 & 19 & 62 & 77 & 1 & \textbf{1232}\tabularnewline
\end{tabular}
\caption{LR}
\end{subtable}
\end{adjustbox}
\caption{Confusion Matrices of all the classifiers $n=64$}
\label{confmats}
\end{table*}

\end{document}